\documentclass[lettersize,journal]{IEEEtran}
\usepackage{amsmath,amsfonts}
\usepackage{array}
\usepackage[caption=false,font=normalsize,labelfont=sf,textfont=sf]{subfig}
\usepackage{textcomp}
\usepackage{stfloats}
\usepackage{url}
\usepackage{verbatim}
\usepackage{graphicx}
\usepackage{cite}
\usepackage{hyperref}
\usepackage{tabu}                     
\usepackage{multirow}                 
\usepackage{multicol}                 
\usepackage{multirow}                
\usepackage{float}                    
\usepackage{makecell}                 
\usepackage{booktabs}                 
\usepackage{stfloats}
\usepackage{color} 
\usepackage{tabularx}
\usepackage[ruled,linesnumbered]{algorithm2e}

\begin{document}

\title{A Real-time Degeneracy Sensing and Compensation Method for Enhanced LiDAR SLAM}

\author{Zongbo Liao$^{1}$, Xuanxuan Zhang$^{1}$, Tianxiang Zhang$^{1}$, Zhi Li$^{2}$, Zhenqi Zheng$^{1}$,\\ Zhichao Wen$^{1}$, and You Li$^{1}$,~\IEEEmembership{ Senior Member, ~IEEE}
\thanks{$^{1}$Zongbo Liao, Xuanxuan Zhang, Tianxiang Zhang, Zhenqi Zheng, and You Li are with the State Key Laboratory of Surveying, Mapping and Remote Sensing, Hubei Luojia Laborotary, Wuhan University, Wuhan 430072, China. \{liaozb, xuanxuanzhang, cyberkona, zhengzhenqi, zhichaowen,  liyou\}@whu.edu.cn. $^{2}$Zhi Li is with the Beijing Automation Control Equipment Research Institute, the CASIC, Beijing, China. lizhi181@mails.ucas.ac.cn. Corresponding authors: Xuanxuan Zhang. This work was supported in part by the National Key R\&D Program of China (2023YFB3906600, 2022YFB3903800, 2022YFE0139300), the National Natural Science Foundation of China (42274052), and the Major Science and Technology Projects in Hubei Province (2022AAA009).}
}



\maketitle

\begin{abstract}
LiDAR is widely used in Simultaneous Localization and Mapping (SLAM) and autonomous driving. The LiDAR odometry is of great importance in multi-sensor fusion. However, in some unstructured environments, the point cloud registration cannot constrain the poses of the LiDAR due to its sparse geometric features, which leads to the degeneracy of multi-sensor fusion accuracy. To address this problem, we propose a novel real-time approach to sense and compensate for the degeneracy of LiDAR. Firstly, this paper introduces the degeneracy factor with clear meaning, which can measure the degeneracy of LiDAR. Then, the Density-Based Spatial Clustering of Applications with Noise (DBSCAN) clustering method adaptively perceives the degeneracy with better environmental generalization. Finally, the degeneracy perception results are utilized to fuse LiDAR and IMU, thus effectively resisting degeneracy effects. Experiments on our dataset show the method's high accuracy and robustness and validate our algorithm's adaptability to different environments and LiDAR scanning modalities. 
\end{abstract}

\begin{IEEEkeywords}
Degeneracy, DBSCAN, LiDAR, SLAM.
\end{IEEEkeywords}

\section{Introduction}
\IEEEPARstart{W}{ith} the development of technologies such as artificial intelligence and SLAM, the field of autonomous driving vehicles has achieved impressive results\cite{chiang2023performance}, \cite{lou2022review}. LiDAR is a research hotspot in the SLAM algorithm because it can directly sense the 3D structure information of the environment and is not easily affected by weather conditions and light \cite{r5}, \cite{r6}. However, the localization based on LiDAR is strongly dependent on the environment's geometric features \cite{r7}. The LiDAR odometry is easily affected when the vehicle moves rapidly or in degenerate geometric structures. In degenerate geometry environments, LiDAR scan-matching algorithms should be more adaptable \cite{r8}. The situation makes the deviations in state estimation and brings great challenges to LiDAR SLAM algorithms. 
{The degenerate environment referred to in this paper is where the lack of point cloud constraints leads to significantly increased uncertainty in state estimation. Mainly, it refers to the lack of geometric structural features; in such environments, the point clouds do not provide uniform constraints in all directions.} 
To alleviate such problems, most studies have been carried out by actively exploring LiDAR degeneracy or adding more constraint information \cite{r9} - \cite{r11}. This paper also inherits this idea and approach. {However, there are still three limitations: first, most research methods are based on the eigenvalues of the optimization problem, whose meaning is unclear, and the metric is difficult to measure \cite{r8}; {Second, determining LiDAR degeneration relies on empirical thresholds. These thresholds vary with the environment and have limited generalization across environments and sensor types;} Third, the robustness of positioning and mapping is improved by adding more sensors or adding some motion constraint information, but it also causes the system to be bloated and more expensive. {Motion constraints require the vehicle to follow specific behaviors, such as zigzag motion \cite{r9}. This restricts their application in diverse environments and motion patterns.}
To address the issues mentioned above regarding the potential degeneracy of LiDAR performance during motion, we propose a novel approach for real-time sensing and resistance of LiDAR degeneracy.} The LiDAR degeneracy perception algorithm based on based on the feature method for LiDAR SLAM is designed in this work. Firstly, we calculate the condition number corresponding to the rotation and translation of the LiDAR, which serves as the degeneracy factor. Then, we introduce the DBSCAN clustering method to determine if the LiDAR observations have degenerated. This degeneracy detection method based on DBSCAN does not need to readjust the degeneracy threshold for different environments and types of LiDAR. The main contributions of this paper are:

\begin{enumerate}
\item{{A degeneracy factor for LiDAR systems has been defined. This method is straightforward to implement, with a clear meaning, and can quantify the degeneracy in LiDAR systems, avoiding the issue of ambiguous significance common to most current degeneracy metrics.}}
\item{{A LiDAR degeneracy perception method based on DBSCAN clustering has been proposed. This method does not rely on empirical thresholds and is not affected by external environments or types of LiDAR. It can robustly detect LiDAR degeneracy.}}
\item{{A method to resist LiDAR degeneracy has been introduced. Based on the results of degeneracy perception and the degeneracy factor, the pose from the IMU are fused with those from the degraded LiDAR through weighted integration for compensation.}}
\end{enumerate}
\begin{figure*}[htb]
\centering
\includegraphics[page=1,width=0.98\textwidth]{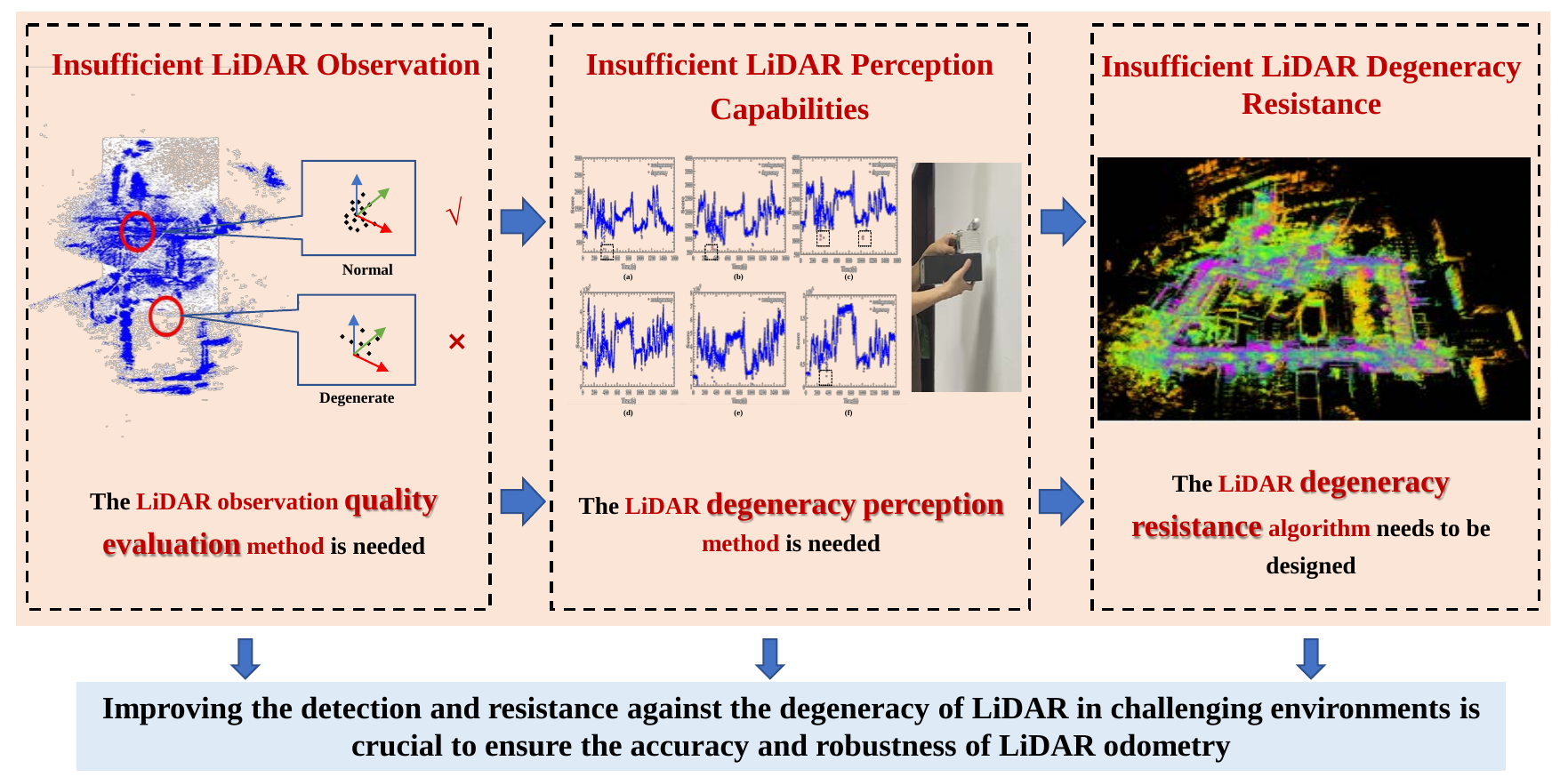}%
\caption{Research background. The lack of adaptability of LiDAR to environments with degenerate geometries leads to errors in state estimation and challenges LiDAR SLAM. The current algorithms use metrics with unclear meaning to determine whether the LiDAR is degenerate, which could be more effective in measuring the degeneracy degree of LiDAR and solving the degeneracy problem of LiDAR.}
\label{fig1}
\end{figure*}

\section{Relater Work}
 Given its critical impact on accurate mapping and localization, developing effective methods to combat degeneracy has become a prominent focus in the field. The current techniques can be broadly classified into two categories:(1)Limit degeneracy by adding more information to the system. (2)Actively deal with degeneracy in SLAM systems by analyzing the constraints imposed by the measurements, identifying degeneracy, and compensating for them.

 Tardioli D \cite{r9} added motion constraints to the LiDAR odometry by peculiar zig-zag movement patterns to limit the degeneracy of the system in a tunnel-like environment. Khattak S \cite{r12} fused visual-inertial odometry (VIO) and thermal-inertial odometry (TIO) estimation with LiDAR odometry (LO); LVI-SAM \cite{r13} and R3LIVE \cite{r14} are the state-of-the-art method that combines LiDAR-inertial odometry (LIO) with visual-inertial odometry (VIO) systems using a tightly coupled framework; LOCUS \cite{r15} constructed a multi-sensor fusion framework around LiDAR to improve the robustness of the system in the degenerate scene. However, the integration of additional sensors into our system poses particular challenges. One of the primary concerns is the increased costs associated with incorporating multiple sensors, which can significantly impact the affordability and feasibility of the overall system, such as synchronization problems and payload limitations \cite{r16}. While effective, this approach limits costs, equipment, and scene requirements.
 
The methods for proactively identifying and detecting degeneracy typically do not add extra sensors, which greatly reduces the cost and system load.
M. Koizumi et al. \cite{koizumi2017avoidance}utilized the Fisher information matrix to calculate the covariance matrix of the positioning error. The largest eigenvalue of the covariance matrix was used as an indicator of the reliability of matching positioning on the map. When the eigenvalue becomes particularly large, it is considered to have entered a degenerate area, indicating that the LiDAR has degenerated. However, this is a LiDAR degeneracy detection scheme based on a priori maps. Zhang J et al. \cite{r10} evaluate degeneracy by calculating the associated eigenvalues and eigenvectors to determine the degeneracy directions and prevent optimization in those directions, inspiring our method’s development. Hinduja A \cite{r21} proposed a degeneracy-aware ICP algorithm based on Zhang J's method. Most of the methods in this category have directly chosen eigenvalues to determine degeneracy.
Zheng et al. \cite{rong2016detection} represented the degeneracy of LiDAR by using the eigenvalues of the Empirical Observability Gramian (EGO) and calculated its local condition number in real-time as a threshold to determine the degeneracy condition. These methods that detect LiDAR degeneracy through eigenvalues have certain limitations because the eigenvalue indicator is difficult to use to measure the degree of LiDAR degeneracy \cite{r8}. Moreover, eigenvalues are quite dependent on environmental variables, such as measurement noise or the number of valid points during point cloud registration, and are highly susceptible to the influence of external environmental factors.

The ratio of eigenvalues can reduce the impact of the environment on eigenvalues. Cho H G et al. \cite{r18} detected degeneracy between two sets of planar features by evaluating the ratio of the eigenvalues of the second-moment matrix. Then, they projected the values of the directions recorded by the IMU onto the directions estimated to degenerate for resistance. Ebadi et al. \cite{ebadi2021dare} proposed to use the condition number of the Hessian matrix as an assessment of the degree of LiDAR degradation. They eliminated the dependence of eigenvalues on environmental variables by calculating the eigenvalue ratio. However, they did not consider the differences in scale and type between the rotation and translation subspaces, which could be important in practical applications.



{Our method also uses the ratio of eigenvalues to detect LiDAR degeneracy. However, we account for the differences in scale and type between rotation and translation.} In addition, most current degeneracy detection methods are highly dependent on the setting of thresholds. However, the setting of thresholds is highly dependent on the environment and the type of LiDAR, which is difficult to predict reliably in advance. We also propose a LiDAR adaptive degeneracy detection algorithm based on DBSCAN, which avoids the issue of adjusting thresholds in different scenarios.

Finally, after detecting the LiDAR degeneracy, the measurement information from the LiDAR odometry and IMU is fully fused to optimize the integration of IMU and LiDAR translation or rotation. IMU is not affected by external environmental conditions, allowing it to autonomously maintain continuous high-frequency positioning \cite{r24}, and the estimation remains stable within a short period of time. Therefore, IMU and LiDAR-based fusion has been widely used. Depending on the design scheme, these methods can be categorized into two main groups: tightly-coupled methods and loosely-coupled methods. For the loosely-coupled method, IMU provides the initial estimate for LiDAR scan registration in LOAM \cite{r25}. Zhen et al. use the error-state EKF to fuse the IMU measurements and pose estimates obtained from the Gaussian particle filter of LiDAR measurements \cite{r16}. For tightly-coupled, there are two main approaches: optimization-based and filter-based. LIOM \cite{r26} presents a graph-optimization-based LiDAR inertial fusion method where the IMU pre-integration is introduced into the odometry. LIO-SAM \cite{r27} uses similar graph-optimization with IMU pre-integration constraints but requires a nine-axis IMU to produce attitude measurement as the prior. For the filter-based approach, LINS \cite{r28} firstly introduces the tightly coupled iterated Kalman filter into LiDAR inertial fusion method to solve the 6DOF ego-motion. FAST-LIO \cite{r29} adopts the same approach but proposes a new formula of the Kalman gain computation to lower the computation load.  {However, these LiDAR-Inertial Odometry (LIO) systems rely heavily on LiDAR-driven technology, often overlooking the synergy between LiDAR and IMU. When the LiDAR performance deteriorates, the overall positioning accuracy of the odometry system can significantly decline, leading to severe drift issues because the valuable information from the IMU cannot be fully leveraged.To address this challenge, we propose utilizing the LiDAR degeneracy factor to make a trade-off between LiDAR and IMU measurements. Thus, robust positioning can be achieved even in LiDAR degeneracy.}

\section{Syetem Overview}
\begin{figure*}[htb]
\centering
\includegraphics[page=2,width=0.98\textwidth]{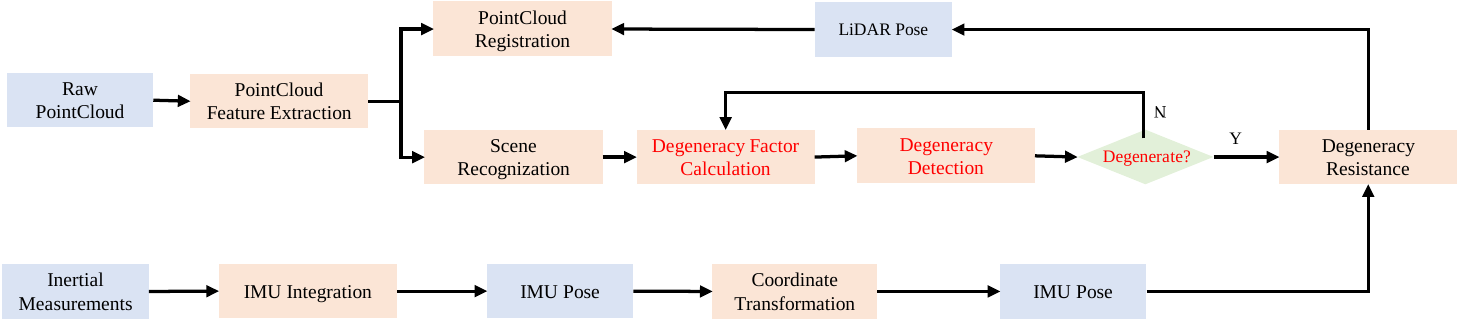}%
\caption{Block diagram illustrating the full pipeline of the proposed Degeneracy-Aware and Resistance method. The system starts with the measurements of LiDAR and IMU. The features are extracted from the point cloud to provide information for later point cloud registration. During the point cloud registration, the LiDAR's degeneracy will be sensed. Simultaneously, the IMU integration will be performed. After LiDAR degeneracy, information from the LiDAR odometry is fused with information from the IMU to resist degeneracy in positioning and mapping accuracy caused by LiDAR degeneracy.} 
\label{fig2}
\end{figure*}

Fig.\ref{fig2} shows the overall framework of the proposed degeneracy-aware method, which consists of two main modules: A degeneracy-aware method based on condition number and DBSCAN and a degeneracy motion state resistance method. These two main modules can be summarized as follows:
\begin{enumerate}
\item{Degeneracy-aware method based on condition number and DBSCAN. During point cloud registration, two matrices related only to rotation and translation are obtained from the Hessian matrix of the distance loss function from the point to the plane. The condition numbers of these two matrices are then calculated, which serve as the degeneracy factor. Then, the DBSCAN-based clustering method senses the degeneracy of LiDAR rotation and translation.}
\item{Degeneracy resistance. When a degeneracy is detected, the rotation or translation state obtained from IMU is projected onto the rotation or translation of LiDAR. By fusing the state estimation of LiDAR and IMU, the method resists environmental degeneracy and improves the robustness of LiDAR localization and mapping.}
\end{enumerate}

\section{Methodology}
In this section, we describe the degeneracy-aware algorithm based on the condition number in detail and propose a novel degeneracy factor with clear meaning. At the same time, the DBSCAN clustering method is used to perceive the degeneracy. When the LiDAR suddenly degenerates, the degeneracy factor will suddenly become more prominent. Compared with that in the normal state of the LiDAR, the degeneracy factor at this time is abnormal. After sensing the LiDAR degeneracy, the motion state of the IMU is used to compensate for the degenerate motion state of the LiDAR.
\subsection{Degeneracy Factor}

{Referring to LOAM \cite{r25}, we construct point-to-surface and point-to-edge distance loss functions.}
\begin{equation}
    f_{\mathcal{E}}\left(\boldsymbol{T}_{k}\right)=d_{\mathcal{E}} 
    \label{con:eqE}
\end{equation}
\begin{equation}
    f_{\mathcal{H}}\left(\boldsymbol{T}_{k}\right)=d_{\mathcal{H}}
    \label{con:eqH}
\end{equation}
{where  $\boldsymbol{T}_{k}$ represents the pose of the LiDAR at time $k$. $ f_{\mathcal{E}}$ denotes the point-to-edge distance function, and $d_{\mathcal{E}}$ denotes the point-to-edge distance residual. $f_{\mathcal{H}}$ denotes the point-to-surface distance function, and $d_{\mathcal{H}}$ denotes the  point-to-surface distance residual.}

{Unify (\ref{con:eqE}) and (\ref{con:eqH}) into the form shown in Eq. (\ref{con:eqZ}), and then superimpose the residual functions of the line-plane distances for all line feature points and plane feature points in the $k$-th frame, to construct the least squares problem shown in Eq. (\ref{con:xxr}). The GaussNewton method is used to optimize.}
\begin{equation}
    f_i\left(\boldsymbol{T}_{k}\right)=d_i,\  \  \boldsymbol{T}_{k}=[\mathbf{R}_{k},\boldsymbol{t}_{k}]
    \label{con:eqZ}
\end{equation}
\begin{equation}
  \label{con:xxr}
  \boldsymbol{T}_{k}^{{*}} =\underset{\boldsymbol{T}_{k}}{\arg \min }\sum_{i=1}^N\left\|\boldsymbol{f}_i(\boldsymbol{T}_{k})\right\|^{2}
\end{equation}
{In this paper, the Gaussian-Newton method\cite{r30} is used to solve the nonlinear least squares problems.}
\begin{equation}
  \label{equ7}
    \sum_{i=1}^N \mathbf{J}_i(\boldsymbol{T}_{k}) f_i(\boldsymbol{T}_{k})+\underbrace{\sum_{i=1}^N \mathbf{J}_i(\boldsymbol{T}_{k}) \mathbf{J}_i^{\top}}_{\mathbf{H}(\boldsymbol{T}_{k})}(\boldsymbol{T}_{k}) \Delta \boldsymbol{T}_{k}=0
\end{equation}
{where $\Delta \boldsymbol{T}_{k}$ is a small perturbation, $\mathbf{J}_i(\boldsymbol{T}_{k})$ and $\mathbf{H}(\boldsymbol{T}_{k})$ are the Jacobian matrix and Hessian matrix of the error function $\boldsymbol{f}_{i}(\boldsymbol{T}_{k})$ respectively. }
\begin{equation}
  \label{equ8}
\mathbf{J}_i(\boldsymbol{T}_{k})=\frac{\partial \boldsymbol{f}_{i}(\boldsymbol{T}_{k})}{\partial \boldsymbol{T}_{k}}=[\frac{\partial \boldsymbol{f}_{i}(\boldsymbol{T}_{k})}{\partial \mathbf{R}}, \frac{\partial \boldsymbol{f}_{i}(\boldsymbol{T}_{k})}{\partial \boldsymbol{t}}]
\end{equation}


{Then, we analyze the causes of the degradation situation based on the above pose estimation process. The pose estimation method used in this paper mainly relies on the correspondence constraint between the feature points in the source point cloud and the corresponding edge and surface features in the target point cloud, i.e., the distance between the feature points and the corresponding edge and surface features tends to be close to zero. For any vehicle motion, there is always a parameter of the pose, which should satisfy:}
\begin{equation}
  \label{equ11}
  {\min }\sum_{i=1}^N\left\|\boldsymbol{f}_i(\boldsymbol{T}_{k})\right\|^{2} \rightarrow  0.
\end{equation}

\begin{figure}[htb]
\centering
\includegraphics[page=1,width=3.5in]{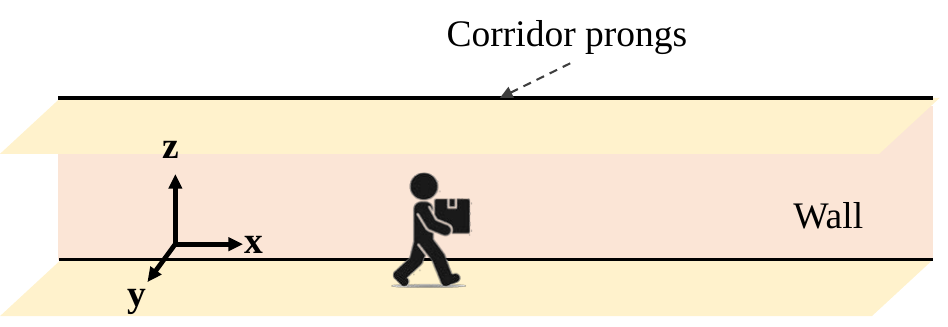}
\caption{Degenerate state Graph. LiDAR is now in a single-degree degeneracy environment.}
\label{fig3}
\end{figure}

{The challenges in the point cloud alignment process become particularly pronounced in environments such as tunnels or long corridors, where LiDAR encounters walls that are symmetric and extend over long distances. These settings are marked by their repetitive geometric structures, lacking geometric features. This can lead to the alignment process yielding multiple potential solutions, thereby impacting the accuracy of motion estimation. {As shown in Fig.\ref{fig3}, pedestrians in a long corridor move straight in the $x$ direction. The corridor is oriented parallel to the travel direction, while the plane normals are perpendicular to it.} In this case, the five degrees of freedom $y,z, roll, pitch, yaw$ are well constrained and the alignment algorithm can accurately estimate these parameters by minimizing the cost function. {However, the situation differs along the $x$-axis, the pedestrian's forward direction.  Feature duplication and lack of directional variation, caused by the geometric layout of long corridors, result in insufficient constraints in the $x$-direction.  This leads to multiple possible solutions along the $x$-axis, making it difficult for the alignment process to determine the correct estimate of $x$.}}

From (\ref{equ7}), the Hessian matrix $\mathbf{H}$ can be obtained:
 \begin{equation}
    \begin{aligned}
        \mathbf{H}
        &=\sum_{i=1}^n \mathbf{J}_i(\boldsymbol{T}) \mathbf{J}_i^{\top}(\boldsymbol{T})\\
        &=\left[\begin{array}{lll}
            \mathbf{J}_{1 r} & \ldots & \mathbf{J}_{n r} \\
            \mathbf{J}_{1 {t}} & \ldots & \mathbf{J}_{n {t}}
            \end{array}\right]\left[\begin{array}{ccc}
            \mathbf{J}_{1 r} & \mathbf{J}_{1 {t}} \\\vdots & \vdots\\
            \mathbf{J}_{n r} & \mathbf{J}_{n {t}}
            \end{array}\right]\\
        &=\left[\begin{array}{ll}
            \mathbf{H}_{r r} & \mathbf{H}_{r t} \\
            \mathbf{H}_{t r} & \mathbf{H}_{t t}
            \end{array}\right]_{6 \times 6.}
    \end{aligned}
 \end{equation}

{The Hessian matrix $\mathbf{H}$ is divided into four sub-matrices according to the relationship between $\mathbf{R}$ and $ \boldsymbol{t}$. The $ \mathbf{H}_{r r}$ sub-matrix only contains information related to  $ \mathbf{R}$, and it can be used to analyze the degradation in the rotation. The $\mathbf{H}_{t t}$ submatrix contains only information related to $\boldsymbol{t}$ and is used to analyze the degeneracy in the translation. The $ \mathbf{H}_{rt}$ and $ \mathbf{H}_{tr}$ sub-matrices represent the interaction information between $\mathbf{R}$ and $ \boldsymbol{t}$, respectively.}

{Considering the differences in scale and type between rotation and translation,  we use only $\mathbf{H}_{r r}$ and $\mathbf{H}_{t t}$ in the eigenanalysis of the SVD. } 
\begin{equation}
    \mathbf{H}_{t t}=\boldsymbol{V}_t \Sigma_t \boldsymbol{V}_t^{\top}=\boldsymbol{V}_t\left[\begin{array}{lll}
          \lambda_{t1}  & 0 & 0 \\
            0 & \lambda_{t2} & 0\\
           0 & 0 & \lambda_{t3} 
            \end{array}\right] \boldsymbol{V}_t^{\top}
\end{equation}

\begin{equation}
    \mathbf{H}_{r r}=\boldsymbol{V}_r \Sigma_r \mathbf{V}_r^{\top}=\boldsymbol{V}_r 
    \left[\begin{array}{lll}
          \lambda_{r1}  & 0 & 0 \\
            0 & \lambda_{r2} & 0\\
           0 & 0 & \lambda_{r3} 
            \end{array}\right]
    \mathbf{V}_r^{\top}
\end{equation}
{where $\Sigma_r$ and $\Sigma_t$ are matrices consisting of the eigenvalues of $\mathbf{H}_{r r}$ and $\mathbf{H}_{t t}$, and $ {\lambda}_1 > {\lambda}_2 > {\lambda}_3$. Then, use (\ref{equ15}) and (\ref{equ16}) as the degeneracy factors for LiDAR rotation and translation, respectively.}
\begin{equation}
  \label{equ15}
  S_{s \mathbf{R}}=\frac{{\lambda}_{r1}}{{\lambda}_{r3}}
\end{equation}
\begin{equation}
  \label{equ16}
  S_{s \boldsymbol{t}}=\frac{{\lambda}_{t1}}{{\lambda}_{t3}}
\end{equation}

{$S_{s \mathbf{R}}$ is the degeneracy factor for rotation and $S_{s \boldsymbol{t}}$ denotes the degeneracy factor for translation. In this paper, the degeneracy factor is defined by the ratio of the largest and smallest eigenvalues (i.e., the condition number) of the submatrices of the Hessian matrix. The condition number was originally used in numerical analysis to measure the stability or sensitivity of a matrix\cite{r22}. Systems with small condition numbers are considered stable because small errors in the input data do not significantly affect the output, thus ensuring the reliability of the model predictions. On the contrary, systems with larger condition numbers are less stable. {In this paper, we define the degeneracy factor of LiDAR based on the condition number. A larger value of the degeneracy factor indicates that the rotational or translational observability is poorer and the state estimated by LiDAR is more likely to degrade.}}
\subsection{Degeneracy-Aware Algorithm}
This paper analyzes the distribution of the degeneracy factor, when the LiDAR is in a structured environment, the value of the degeneracy factor is relatively small due to the richness of the geometric structural feature information of the environment. On the contrary, when the LiDAR is in an unstructured environment, the scarcity of structural feature information in the environment leads to the degeneracy of the LiDAR, at which time the value of the degeneracy factor increases significantly. In this case, the degeneracy factor of the degeneracy moment relative to the normal moment is the outliers, and by identifying these outliers, we can effectively determine whether the LiDAR is degraded at the current moment. Therefore, DBSCAN clustering is used to detect outliers and further sense the degeneracy of LiDAR automatically.

The DBSCAN algorithm was first proposed by Ester et al. \cite{r32}, and can adaptively recognize dense regions of data and treat data points in sparse regions as noise or outliers. This feature allows DBSCAN to not only handle clusters of arbitrary shapes but also eliminates the need to preset the number of clusters, which greatly enhances the ability to adaptively detect the degraded state of LiDAR. the DBSCAN algorithm is based on two main parameters: $\epsilon$(Eps) and Minpts. Eps defines the range of a point's neighborhood, and MinPts is the minimum number of points required to form a dense region. region; and MinPts is the minimum number of points required to form a dense region. Based on these parameters, DBSCAN classifies data points into three categories (1) Core points, points that have at least MinPts of points within their a-neighborhood. (2)Boundary points, points that are not core but are within the a-neighborhood of a core point. (3) Noise point, a point that is neither a core nor a boundary point.

\begin{figure}[ht]
\centering
\includegraphics[page=4,width=3.6in]{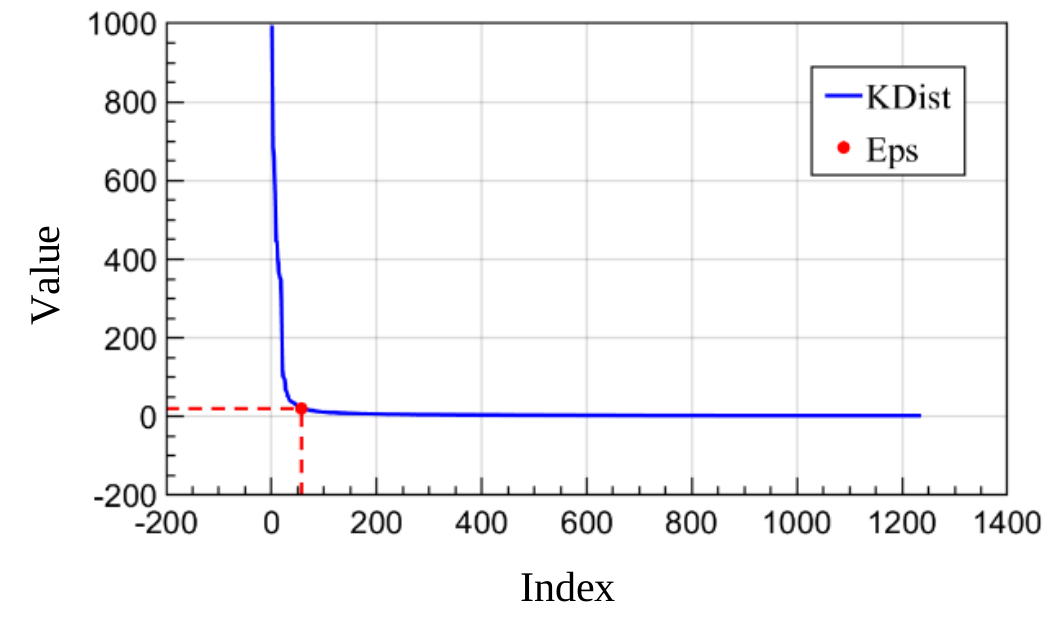}
\caption{K-distance Graph. From a sample data set, we draw the k-distance graph and obtain Eps. The $\mathrm{X}$-axis represents the index of the data, and the $\mathrm{Y}$axis represents the value of the data. Eps is at the inflection point of the k-distance graph.}
\label{fig4}
\end{figure}

{The value of the degeneracy factor is relatively low for LiDAR under normal operating conditions, similar to the core or boundary points in the DBSCAN algorithm. However, when the LiDAR is in a degraded state, the value of the degeneracy factor increases significantly and is equivalent to a noise point in the algorithm. To accurately apply the DBSCAN algorithm to analyze the degeneracy factor, a key step is to correctly set the two parameters of the algorithm: the MinPts and the Eps. Based on the characteristics of the dataset, this paper adopts a common heuristic to adjust the MinPts parameter, i.e., setting the MinPts to $2^d-1$ based on the dimensionality (d) of the data. The data for the degeneracy factor consists of two-dimensional data - time and the numerical magnitude of the degeneracy factor - and thus the choice was made to set MinPts to 3. Further, to determine the optimal value of Eps, this paper employs a $k$-distance graph \cite{rong2016detection} approach. The $k$-distance graph can identify points where there is a significant change in data density, i.e., inflection points where there is a transition from a high-density region to a low-density region. Fig.\ref{fig4} provides an example. The point where the two red dashed lines intersect represents the inflection point, which corresponds to the value of Eps.  Accurately localizing this inflection point can set the optimal Eps value. Our approach is to start traversing from the first point on the $k$-distance graph and focus on the difference in distance from the next three points. If the difference in distance from point $i$ to the next three points is consistently small, it indicates that the change in distance from point $i$ is small. Therefore, point $i$ is likely to be the inflection point, or Eps, that is being searched for. The pseudocode for determining the value of Eps is presented in Algorithm ~\ref{alg:eps}.}

\begin{algorithm}[htb]
 
  \SetAlgoLined
  \KwData{degeneracy facto dataset $D$,  MinPts}
  \KwResult{value of Eps }

  \For{p in $D$}{
    \For{q in $D$}{
        Calculate the distance $di$ between {\textit{p} and \textit{q}}, and add $di$ to a temporary list $A$.
    }
  }
  
{Sort $A$ in ascending order of distance.}

{ Add the MinPts-th smallest distance $A[MinPts]$ to the distances list $B$.}

  Sort $B$ in descending order.
  
  \For{$i \gets 0$ \KwTo $B.size()-3$}{
        \If{B[k] - B[k+1] $\leq $ 0.1 and
        B[k] - B[k+2] $\leq $ 0.1, and
        B[k] - B[k+3] $\leq $ 0.1}{
            Eps = $B[k]$;
            
            break;
        }
    } 
 
  \caption{{Eps Determining.}}
   \label{alg:eps}
\end{algorithm}

The pseudocode of the degeneracy-aware is shown in Algorithm~\ref{alg:dbscan}, where $\mathbf{D}$ is the degeneracy factor data set, $d$ represents the degeneracy factor at the current moment, $x_m$ is the maximum degeneracy factor at normal moments, and $\boldsymbol{y}$ represents whether the LiDAR degenerates at the current moment. {To ensure the efficiency of the algorithm, the size of the data processed by DBSCAN is capped at a maximum of 1000 data points.} In addition, to ensure the accuracy of outlier detection, i.e., when evaluating the degeneracy of LiDAR performance, a constantly changing degeneracy threshold is set. Only when a data point is recognized as noise by the DBSCAN algorithm and exceeds the degeneracy threshold is the LiDAR degraded at the current moment. {The degeneracy checks are only conducted when the dataset comprises 400 points or more. This implies that, if the LiDAR degenerates during this initial phase, our algorithm may fail to accurately detect such degeneracy. This could not only affect the detection at that moment but might also negatively impact subsequent degeneracy assessments.}

\begin{algorithm}[htb]
 
  \SetAlgoLined
  \KwData{degeneracy facto dataset $D$, current degeneracy factor $d$, the maximum degeneracy factor at the normal moment $x_m$}
  \KwResult{ Whether the LiDAR degenerates $y$ }

  \While{$D$.size() $<$ 400}{
    $x_m=max(x_m,d)$
    
    return
  }
   \While{$D$.size() $>$ 1000}{
   { $ pop\_front({D})$}
  }

  Use Algorithm~\ref{alg:eps} to obtain Eps.

  $\textbf{DBSCAN}\ \text{obtains whether}\ d \ \text{is an outlier} $

  \eIf{$d$ is a outlier and $d>x_m$}{
     $y= $ yes
  }{
    $y= $ no

    $x_m=max(x_m,d)$
  }
  \caption{Degeneracy Sensing.}
   \label{alg:dbscan}
\end{algorithm}

\subsection{ Degeneracy Resistance Algorithm}
{This section focuses on mitigating the impact of LiDAR degeneracy in motion states by using LiDAR degeneracy perception and degeneracy factor.} This is achieved by projecting and fusing the state estimation of the IMU onto the direction of LiDAR degeneracy. This approach involves the weighted fusion of motion state data from two different sensors: On one hand, the state estimation derived from point clouds, weighted by degeneracy factors, maintains the constraints from points to lines and surfaces; on the other hand, it capitalizes on the short-term accurate motion state estimation provided by the IMU. In the face of LiDAR degeneracy, this method employs weighted fusion by projecting the IMU's state estimation onto the direction of LiDAR degeneracy, effectively addressing the challenges posed by LiDAR degeneracy.
\begin{equation}
  \label{equ17}
  \begin{aligned}
  \left[\begin{array}{cc}
  \mathbf{R}_{\boldsymbol{L}}^{\prime} & \boldsymbol{t}_{\boldsymbol{L}}^{\prime} \\
  \mathbf{0} & \mathbf{1}
  \end{array}\right] &= \left[\begin{array}{cc}
  \mathbf{R}_{\boldsymbol{b}}^{\boldsymbol{L}} & \boldsymbol{t}_{\boldsymbol{b}}^{\boldsymbol{L}} \\
  \mathbf{0} & \mathbf{1}
  \end{array}\right] \left[\begin{array}{cc}
  \mathbf{R}_{\boldsymbol{b}} & \boldsymbol{t}_{\boldsymbol{b}} \\
  \mathbf{0} & \mathbf{1}
  \end{array}\right] \\
  &= \left[\begin{array}{cc}
  \mathbf{R}_{\boldsymbol{b}}^{\boldsymbol{L}} \mathbf{R}_{\boldsymbol{b}} & \mathbf{R}_{\boldsymbol{b}}^{\boldsymbol{L}} \boldsymbol{t}_{\boldsymbol{b}} + \boldsymbol{t}_{\boldsymbol{b}}^{\boldsymbol{L}} \\
  \mathbf{0} & \mathbf{1}
  \end{array}\right]
  \end{aligned}
\end{equation}

{If $S_{s \boldsymbol{t}}$ or $S_{s \mathbf{R}}$ is abnormal, we consider that the LiDAR is degenerate, so the state estimation of the IMU is projected to the direction of the LiDAR, as shown in (\ref{equ17}), where $\mathbf{R}_{\boldsymbol{b}}$ and $\boldsymbol{t}_{\boldsymbol{b}}$ are rotation and translation in the IMU coordinate system. $\mathbf{R}_{\boldsymbol{L}}^{\prime}$ and $\boldsymbol{t}_{\boldsymbol{L}}^{\prime}$ are the results of the rotation and translation under the IMU coordinate system converted to the LiDAR coordinate system, and $\mathbf{R}_{\boldsymbol{b}}^{\boldsymbol{L}}$ and $\boldsymbol{t}_{\boldsymbol{b}}^{\boldsymbol{L}}$ are the rotation and translation from the IMU to the LiDAR. 
{If degeneracy occurs in translation, the IMU and LiDAR translations are weighted and fused, as shown in (\ref{equ18}). If degeneracy occurs in rotation, the IMU and LiDAR rotations are weighted and fused, as shown in (\ref{equ19}). In these equations, $\mathbf{q}_{\boldsymbol{L}}$ and $\boldsymbol{t}_{\boldsymbol{L}}$ denote rotation and translation in the LiDAR coordinate system, and $\mathbf{q}$ is the quaternion corresponding to the rotation.}}
\begin{equation}
  \label{equ18}
  \boldsymbol{t}=(1-\frac{1}{S_{s \boldsymbol{t}}}) \boldsymbol{t}_{\boldsymbol{L}}^{\prime}+\frac{1}{S_{s \boldsymbol{t}}} \boldsymbol{t}_{\boldsymbol{L}}
\end{equation}
\begin{equation}
  \label{equ19}
  \mathbf{q}=(1-\frac{1}{S_{s \mathbf{R}}}) \mathbf{q}_{\boldsymbol{L}}^{\prime}+\frac{1}{S_{s \mathbf{R}}} \mathbf{q}_{\boldsymbol{L}}
\end{equation}

\section{Experiments}
\begin{figure}[htb]
\centering
\subfloat[]{\includegraphics[page=5, width=1.6in,height=1.8in]{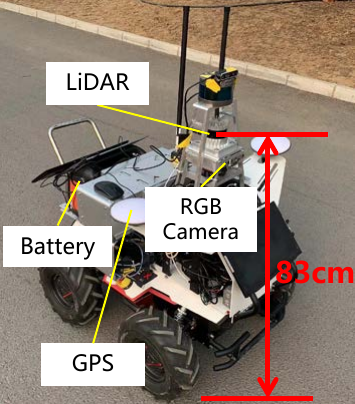}%
\label{fig5a}}
\hspace{0.25cm}
\subfloat[]{\includegraphics[page=6,width=1.6in,height=1.8in]{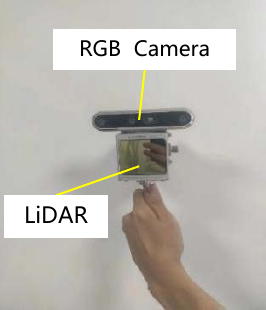}%
\label{fig5b}}
\caption{Hardware system. (a) is a four-wheel robot car with an Ouster-64 LiDAR, GPS, and a Realsense-435i camera. (b) is a handheld LiDAR that includes a Livox-Avia LiDAR and a Realsense-435i camera.}
\label{fig5}
\end{figure}

{The research presented in this paper is validated using solid-state and spinning LiDAR systems. 
In our study, we evaluated our approach using eight different sequences. The sequences xhl, basement, uzh, degenerate\_00, and degenerate\_01 were collected using solid-state LiDAR, while campus, construction, and quad were collected using spinning LiDAR. The sequences degenerate\_00 and degenerate\_01 are from R3LIVE dataset.  The sequences basement, uzh, and construction can be obtained from Hilti SLAM Challenge\cite{r33}. The sequence quad is from Newer College Dataset. The sequences campus and xhl were collected using our equipment, as shown in Fig.\ref{fig5}   These sequences were chosen to evaluate the generalization capability of the degeneracy detection algorithm proposed in this paper across different sensor types. Additionally, these sequences were collected by human operators following pre-planned routes. This paper assumes the sensors have already undergone extrinsic calibration and hardware time synchronization. We also used a soft synchronization method to calibrate the timestamps of the LiDAR and IMU. A brief overview of these 8 sequences is shown in Table \ref{tabel1}.}
\begin{table}[htb]
 \begin{center}
 \caption{{The Brief Overview of Sequences}}
 \label{tabel1}
  \begin{tabular}{cccc}
  \hline
   \multirow{2}{*}{\thead{Sequence}} & \multirow{2}{*}{\thead{Traveling \\ length (m)}} & \multirow{2}{*}{\thead{characteristic}} & \multirow{2}{*}{\thead{Sensor}} \\
   & &  &\\
   & &  &\\
   \hline
   & &  &\\
   campus & 791.5 & severe shaking  & Ouster-64\\
   \hline
   xhl & 275.2 & approaches to the walls  & Livox-Avia \\
   \hline
   basement &  59.1 & approaches to the  walls & Livox-Mid70 \\
   \hline
   uzh & 77.3 & approaches to the walls  &  Livox-Mid70 \\
   \hline
   construction & 36.45 & \thead{aggressive motion, \\approaches to the walls}  &  PandarXT-32 \\
   \hline
   quad & 234.8 & \thead{aggressive motion, \\approaches to the walls}  &  Ouster-64 \\
   \hline
   degenerate\_00 & 75.2 & approaches to the ground  &  Livox-Avia\\
   \hline
   degenerate\_01 & 74.9 & \thead{approaches to the \\ ground and walls}  &  Livox-Avia\\
  \hline
  \end{tabular}
 \end{center}
\end{table}

\begin{figure}[!h]
\centering
\subfloat[]{\includegraphics[page=1, width=1.8in]{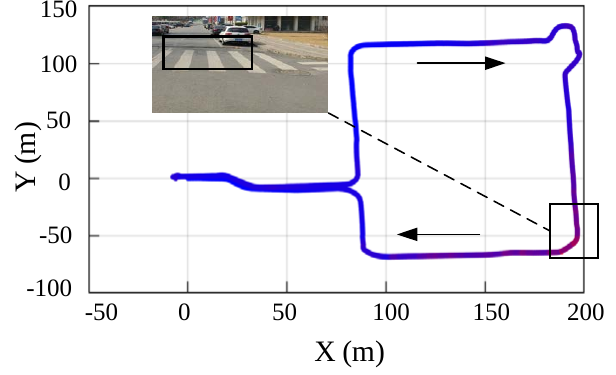}%
\label{fig6a}}
\subfloat[]{\includegraphics[page=2,width=1.8in]{expfig1.pdf}%
\label{fig6b}}
\caption{Degeneracy in campus. (a)The degeneracy varies with the position. (b) The degeneracy changes over time.}
\label{fig6}
\end{figure}

\begin{figure}[!h]
\centering
\subfloat[]{\includegraphics[page=3, width=1.8in]{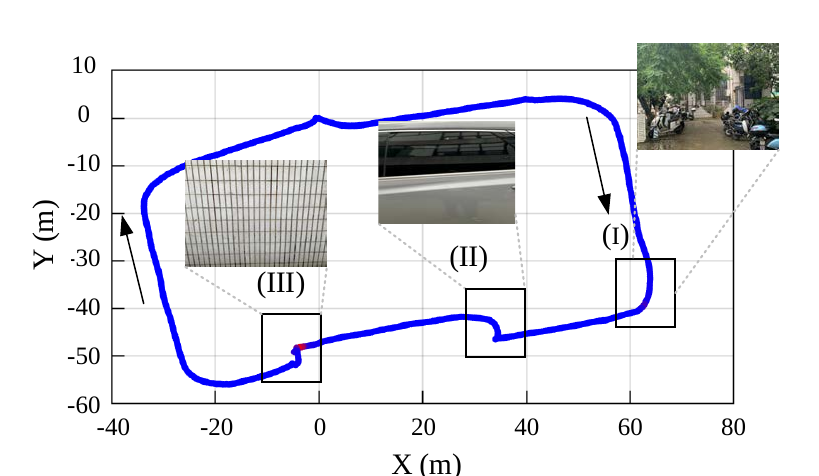}%
\label{fig7a}}
\subfloat[]{\includegraphics[page=4,width=1.8in]{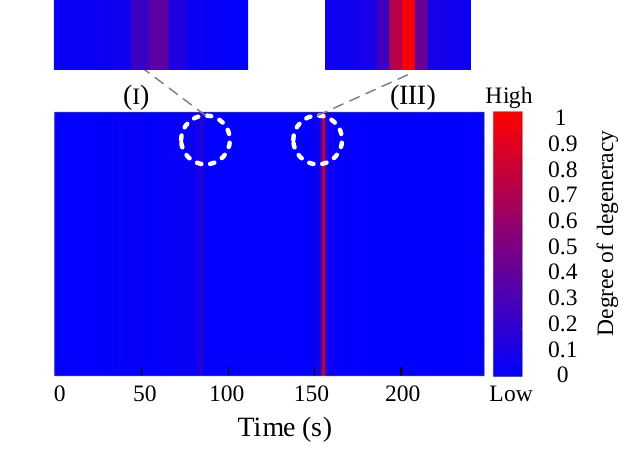}%
\label{fig7b}}
\caption{Degeneracy in xhl. (a)The degeneracy varies with the position. (b) The degeneracy changes over time.}
\label{fig7}
\end{figure}

\begin{figure}[!h]
\centering
\subfloat[]{\includegraphics[page=5, width=1.8in]{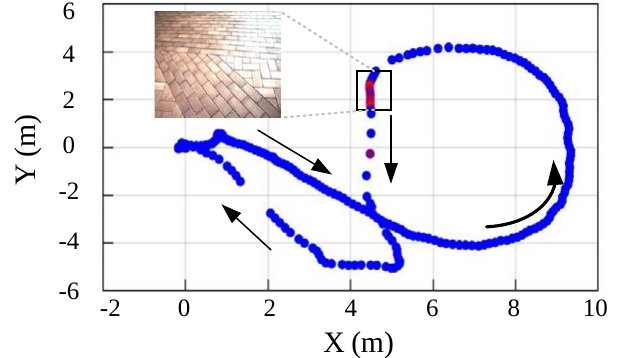}%
\label{fig8a}}
\subfloat[]{\includegraphics[page=6,width=1.8in]{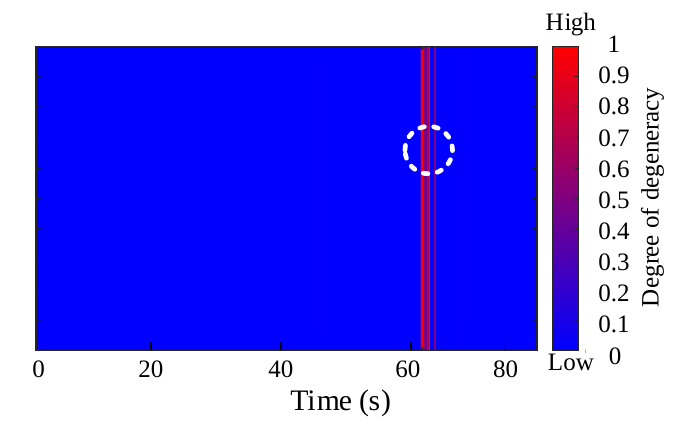}%
\label{fig8b}}
\caption{Degeneracy in  degenerate\_01. (a)The degeneracy varies with the position. (b) The degeneracy changes over time.}
\label{fig8}
\end{figure}

Figs.\ref{fig6}-\ref{fig8} show the degeneracy of campus, xhl, and degenerate\_00. (a) shows that the degeneracy varies with the position. (b) shows that the degeneracy changes over time. (b) is a one-dimensional graph, the horizontal axis represents time, and the vertical axis has no practical meaning. We present the degree of LiDAR degeneracy at each position and time. The color gradient from blue to red signifies the degree of degeneracy, with red indicating higher degeneracy. Consequently, the more intense the red color, the more severe the degeneracy. The degeneracy in the trajectory is framed by a box, which also corresponds to the degeneracy at that moment inside the circle on (b).

{The sequence campus comprises realistic outdoor road environments featuring various elements such as speed bumps and long slopes. The positions of the start and the end of the trajectory overlap. To establish ground truth, we employed commercial GNSS-RTK on the robot, which provides accurate positioning information. In the place marked with the black box in Fig.\ref{fig6a}, this place is a speed bump and also in a curve. The color of the trajectory at this place is red, and in Fig.\ref{fig6b}, the color at that moment is also red. LiDAR degenerates here.}

{The sequence xhl was collected using a handheld device. During the data collection, the device was quickly passed through a curve (Fig. \ref{fig7} (I), degeneracy I), followed by a car (Fig.\ref{fig7} (II), degeneracy II), and then a wall (Fig.\ref{fig7} (III), degeneracy III) that obstructed the LiDAR view. The LiDAR degenerated in these three positions due to the challenging environment. From Fig.\ref{fig7b}, it can be seen that the red at position I is not obvious, where LiDAR end positions only slightly degenerated, and the red at position II is invisible, where LiDAR only undergoes extremely weak degeneracy, but at position III the red is exceptionally bright, where LiDAR suffers severe degeneracy.}

{In degenerate\_00 the LiDAR is scanned toward the ground during the movement, at that time the environment's geometric features are not obvious and the LiDAR undergoes severe degeneracy.}

\begin{figure*}[htb]
\centering
\subfloat[]{\includegraphics[page=7, width=2.4in]{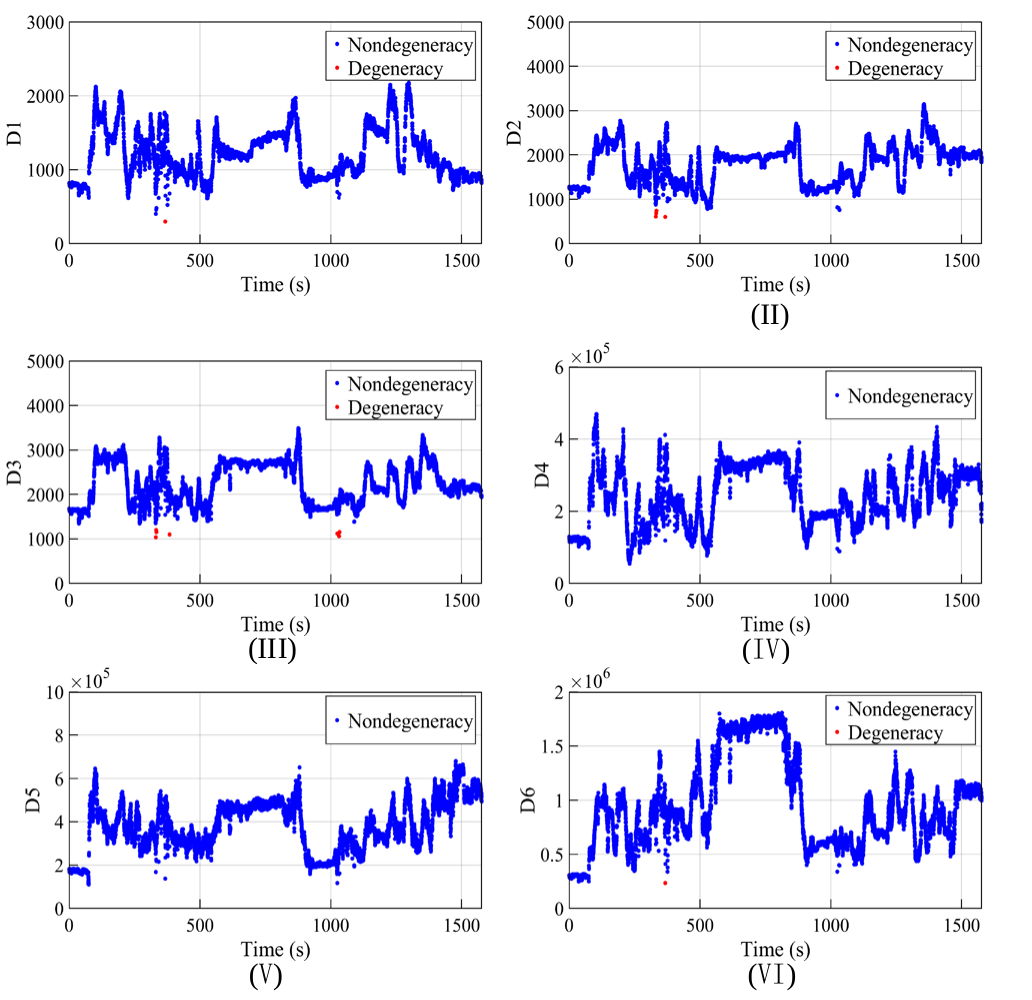}%
\label{fig9a}}
\subfloat[]{\includegraphics[page=8,width=2.4in]{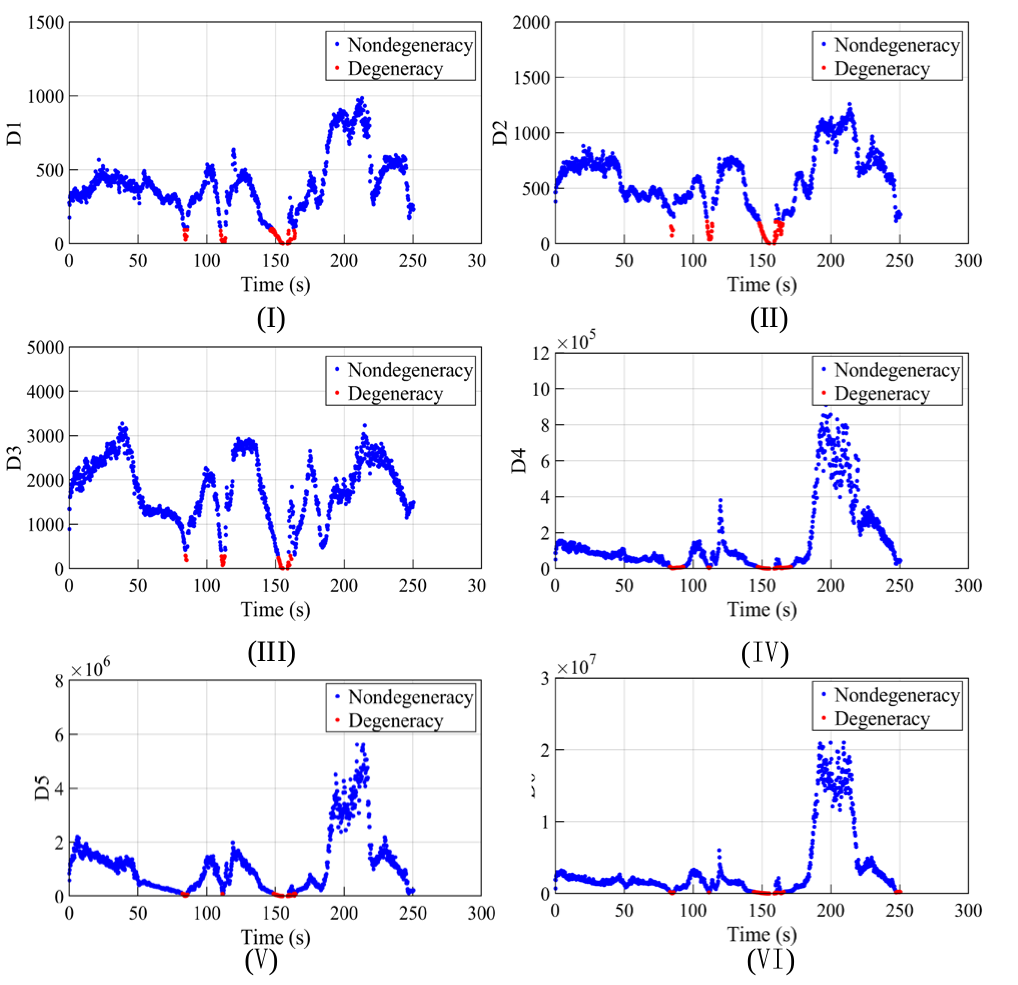}%
\label{fig9b}}
\subfloat[]{\includegraphics[page=9,width=2.4in]{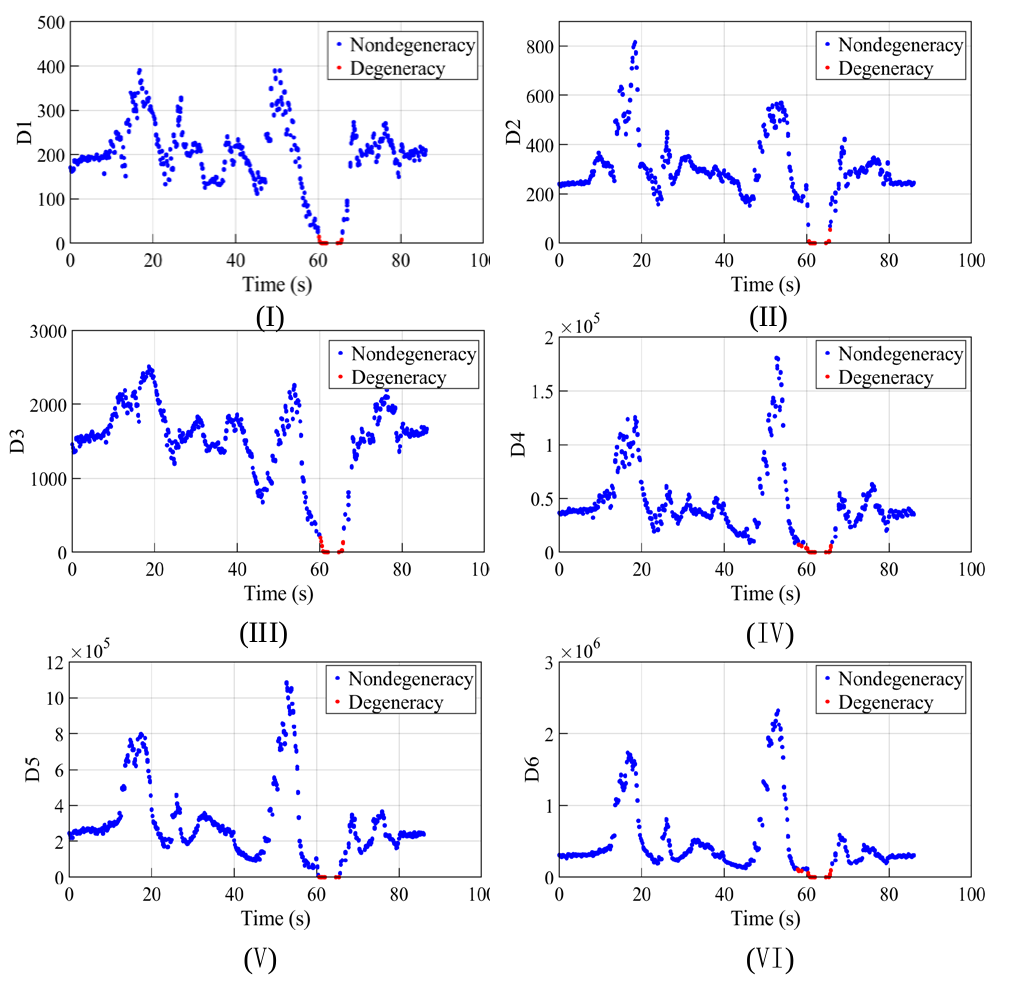}%
\label{fig9c}}
\caption{The degeneracy factor of \cite{r10} on the three sequences. (a), (b) and (c) denote the degeneracy factors of campus, xhl, and degenerate\_00 respectively. Six different degeneracy detection thresholds are set, where degeneracy moments are plotted with red dots. (I)-(VI) denote, respectively, z, y, x, yaw, pitch and yaw.}
\label{fig9}
\end{figure*}


\subsection{Experiments in Degeneracy Sensing}
To verify the performance of the proposed degeneracy-aware method, and to determine whether it can correctly recognize degeneracy, it is compared with the method in \cite{r10}. The method in \cite{r10} uses six eigenvalues to represent six degrees of freedom (x, y, z, roll, pitch, yaw). When the eigenvalue is smaller than the threshold, it is considered that the corresponding degree of freedom has degenerated. However, such a method must set six thresholds, which will differ in different scenes, increasing the difficulty of detecting degeneracy. Our proposed method is essentially two degrees of freedom (rotation and translation), there is no need to set thresholds, and it adaptively perceives degeneracy through the DBSCAN approach.
\begin{figure*}[!h]
\centering
\subfloat[]{\includegraphics[page=13,height=1.5in ,width=2.45in]{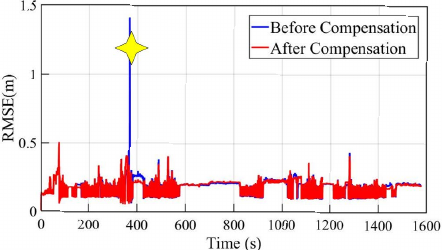}%
\label{fig11a}}
\subfloat[]{\includegraphics[page=14,height=1.6in,width=2.45in]{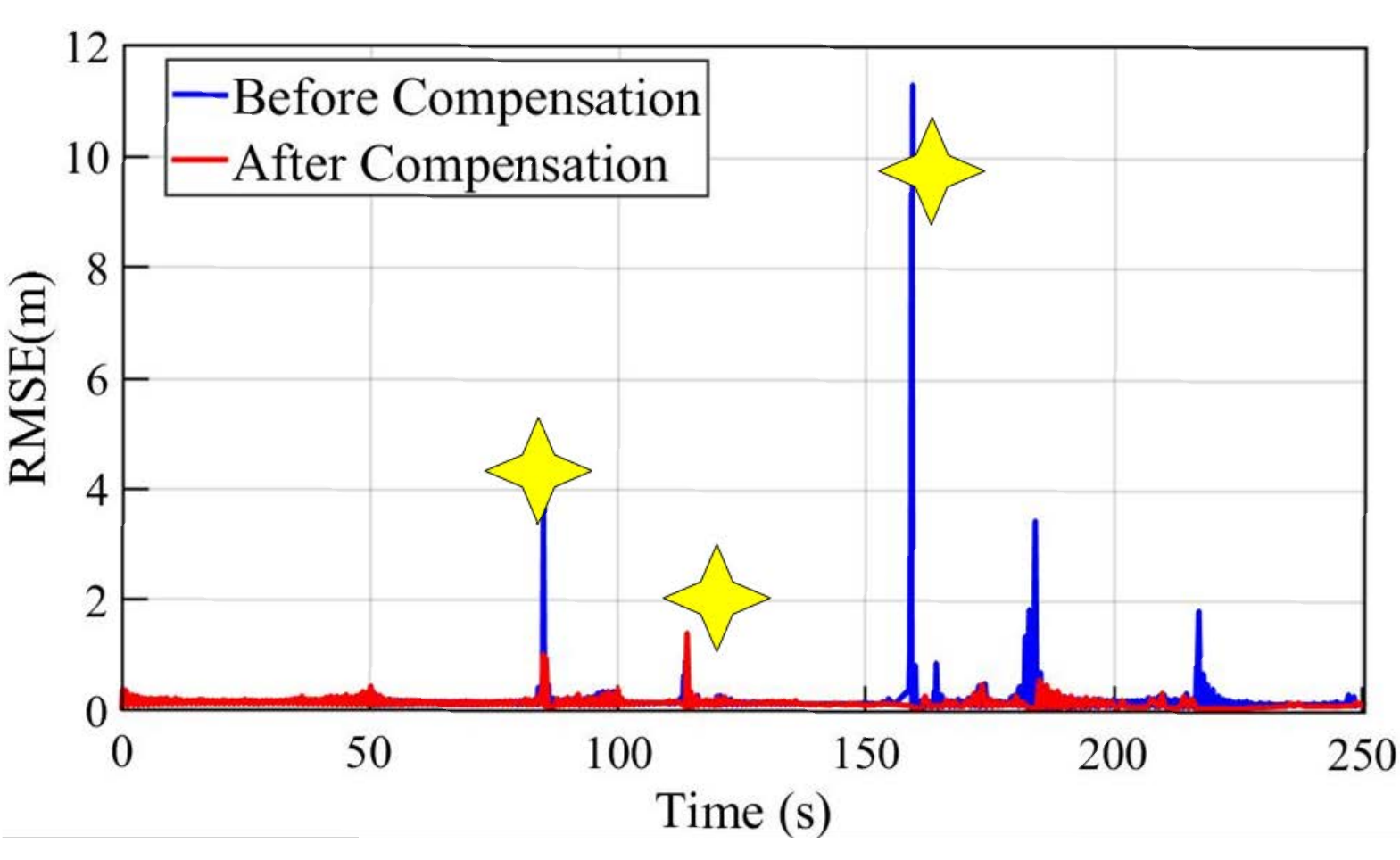}%
\label{fig11b}}
\subfloat[]{\includegraphics[page=15,height=1.6in,width=2.45in]{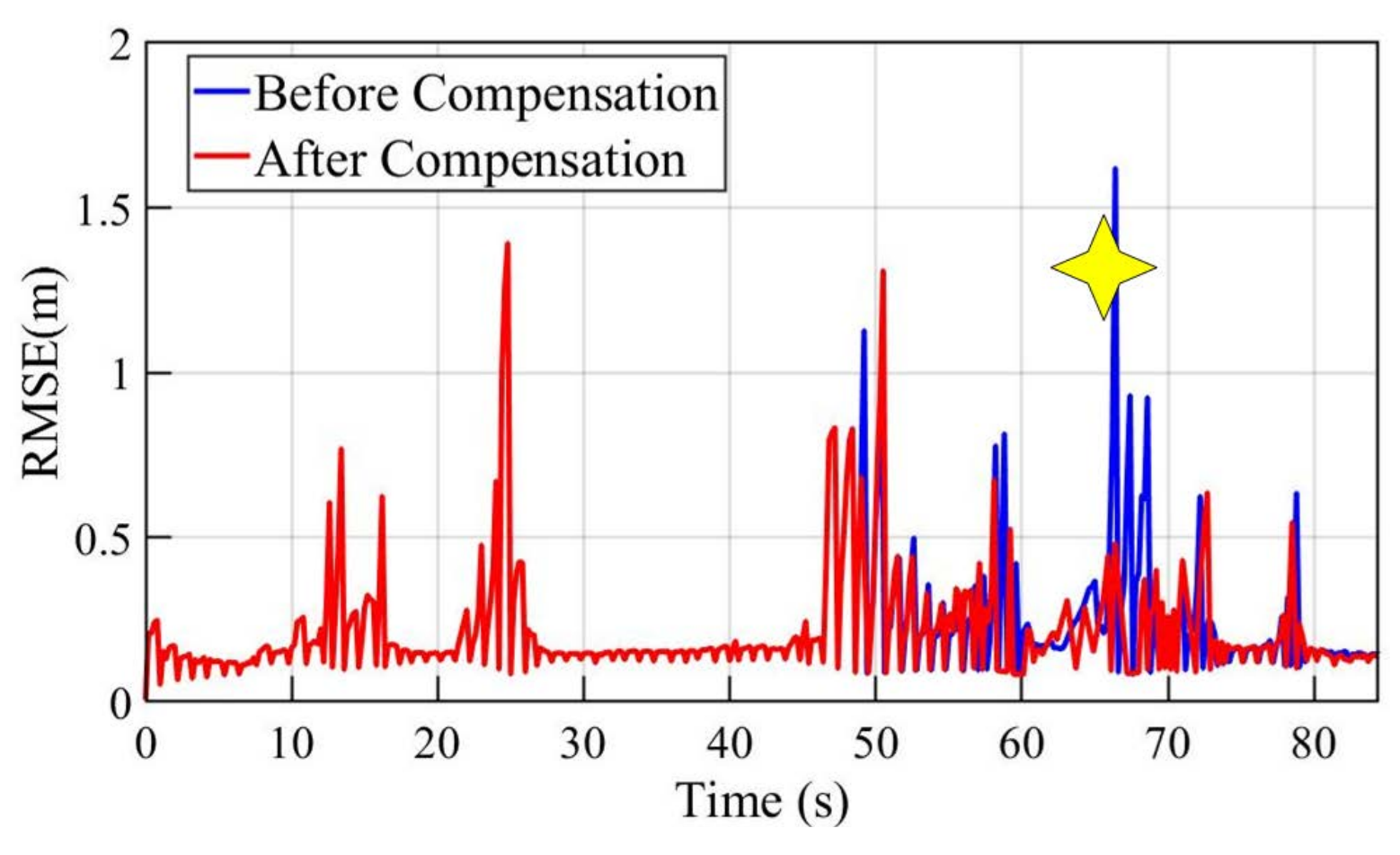}%
\label{fig11c}}
\caption{The RMSE of point cloud registration before and after the degeneracy compensation. (a), (b) and (c) denote the degeneracy factors of campus, xhl, and degenerate\_00 respectively. The yellow four-pointed star in the graph indicates that LiDAR degeneracy occurred during that period.}
\label{fig11}
\end{figure*}
\begin{figure}[!h]
\centering
\subfloat[]{\includegraphics[page=10, width=3.4in]{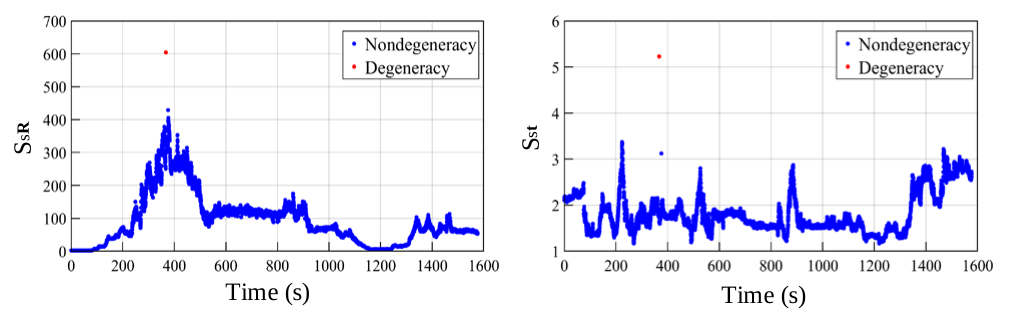}%
\label{fig10a}}\\
\subfloat[]{\includegraphics[page=11,width=3.4in]{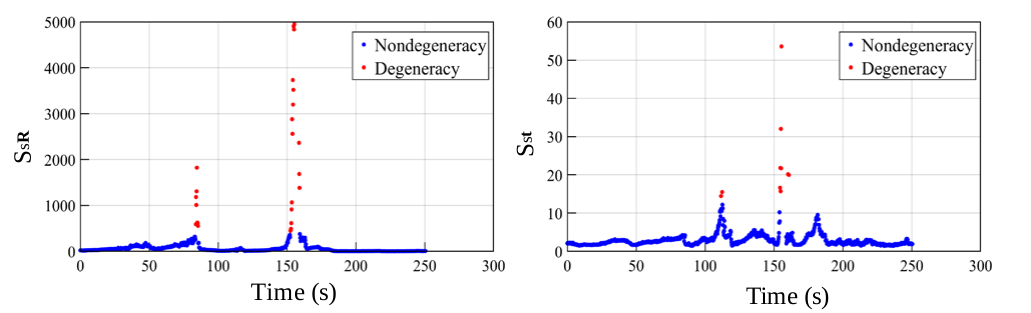}%
\label{fig10b}}\\
\subfloat[]{\includegraphics[page=12,width=3.4in]{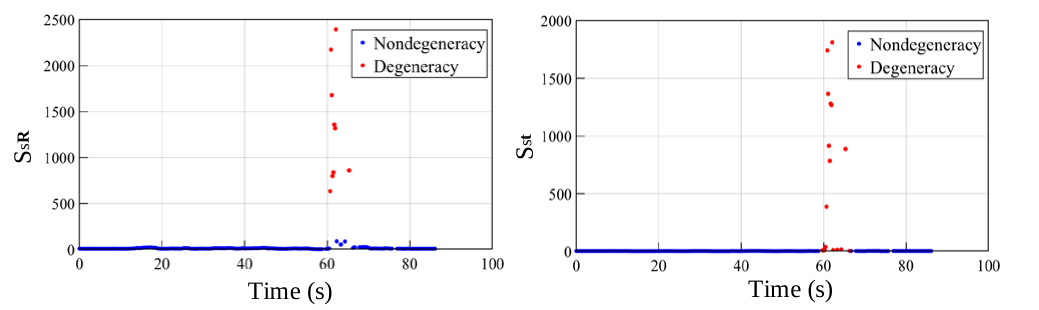}%
\label{fig10c}}
\caption{The degeneracy factor of our method on the three sequences. (a), (b) and (c) denote the degeneracy factors of campus, xhl, and degenerate\_00 respectively. The left shows the degeneracy factor for rotation, and the right shows the translation, where the red dots indicate that degeneracy occurred at that moment.}
\label{fig10}
\end{figure}
\begin{figure*}[htb]
\centering
\includegraphics[page=1,width=0.95\textwidth]{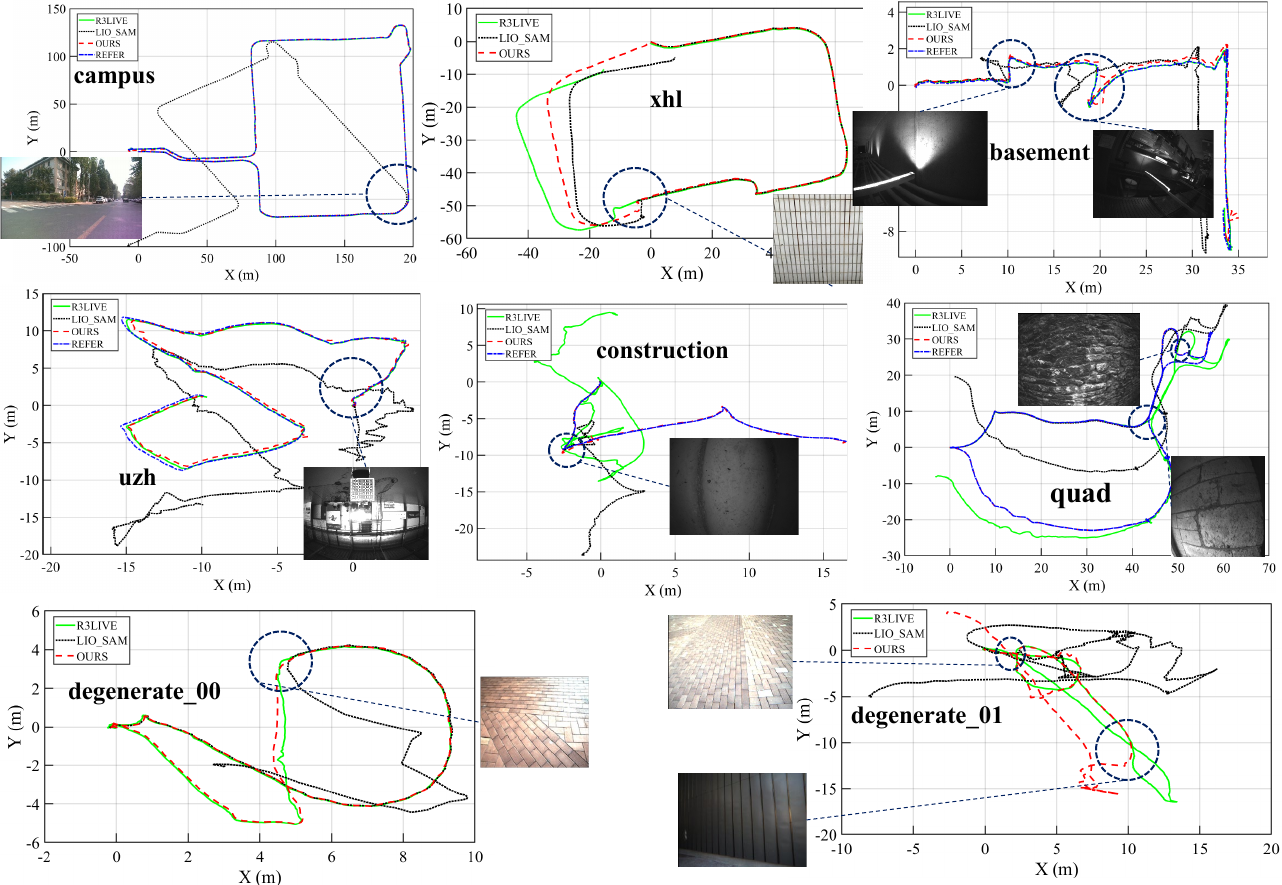}%
\caption{{The trajectory results. The circle above the trace indicates that  LIO-SAM or OURS has drifted at that location, and shows the degenerate environment.}} 
\label{fig12}
\end{figure*}
Fig.\ref{fig9} shows the degeneracy-aware method in \cite{r10}, where the red dot indicates that degeneracy is detected at this moment when an appropriate threshold is set. Fig.\ref{fig10} shows our degeneracy-aware method. Taking the degeneracy factor of xhl as an example, it can be seen from Fig.\ref{fig9b} and Fig.\ref{fig10b} that LiDAR degeneracy occurs at three moments. From Fig.\ref{fig9b}, it can be found that not only is there no significant difference in the magnitude of the values of the degeneracy factors at these three moments, but at the time of both severe and slight degeneracy all six degrees of freedom degenerated at the same time. The degeneracy of these three moments cannot be quantified. However, from Fig.\ref{fig10b}, it is evident that only the rotation degenerates during degeneracy I. This can be attributed to the collected data passing through a curve, leading to the specific degeneracy in rotation while the translation remains almost unaffected. At degeneracy II, only the translation degenerates. This occurs when the LiDAR is close to the car and moving in a straight line, resulting in degeneracy in translation while the rotation remains unaffected. On the other hand, at degeneracy III, both the rotation and translation degenerate. This happens when the LiDAR's view is completely obstructed by the wall, leading to a significant lack of constraints on the rotation and translation of the LiDAR. As a consequence, both rotation and translation experience degeneracy at this point. This also proves the first innovation of this paper, that the degeneracy factor we propose has a clear and strong meaning, and can quantify the degeneracy of LiDAR.

\subsection{Experiments in Degeneracy Resistance}
\renewcommand{\arraystretch}{1.5}
\begin{table}[htb]
\centering
\caption{{Comparative Results of Different Methods($\textbf{no loop-closure}$) on Eight Sequences}}
\label{tb2}
\begin{tabular}{lcccc}
\Xhline{2pt}
\multirow{2}{*}{}& \multirow{2}{*}{Methods} & \multirow{2}{*}{\thead{RMSE(m)}} & \multirow{2}{*}{\thead{End-to-end \\ error(m)}} & \multirow{2}{*}{\thead{Max\\ error(m)}} \\
& & & & \\
\Xhline{1pt}
\multirow{3}{*}{ campus } 
& LIO-SAM & 16.97 & $\backslash$ & 61.03 \\
& R3LIVE & 0.28 & $\backslash$ & 0.83 \\
& OURS & $\mathbf{0.22}$ & $\backslash$ & $\mathbf{0.24}$ \\
\hline 
\multirow{3}{*}{ xhl } 
& LIO-SAM & $\backslash$ & 14.73 & $\backslash$ \\
& R3LIVE & $\backslash$ & 4.81 & $\backslash$ \\
& OURS & $\backslash$ & $\mathbf{0.66}$ & $\backslash$ \\
\hline 
\multirow{3}{*}{ basement } 
& LIO-SAM & 1.04 & $\backslash$ & 2.57 \\
& R3LIVE & $\mathbf{0.1}$ & $\backslash$ & $\mathbf{0.38}$ \\
& OURS & {0.31} & $\backslash$ & {2.15} \\
\hline 
\multirow{3}{*}{ uzh } 
& LIO-SAM & 3.78 & $\backslash$ & 6.58 \\
& R3LIVE & $\mathbf{0.19}$ & $\backslash$ & $\mathbf{0.47}$ \\
& OURS & {0.26} & $\backslash$ & {0.76} \\
\hline
\multirow{3}{*}{construction} 
& LIO-SAM & 10.42 & $\backslash$ & 23.62 \\
& R3LIVE & 10.01 & $\backslash$ & 24.10 \\
& OURS & $\mathbf{0.19}$ & $\backslash$ & $\mathbf{1.01}$ \\
\hline
\multirow{3}{*}{quad } 
& LIO-SAM & 15.79 & $\backslash$ & 28.63 \\
& R3LIVE & 3.28 & $\backslash$ & 9.11 \\
& OURS & $\mathbf{0.14}$ & $\backslash$ & $\mathbf{0.29}$ \\
\hline
\multirow{3}{*}{ degenerate\_00 } 
& LIO-SAM & $\backslash$ & 2.41 & $\backslash$ \\
& R3LIVE & $\backslash$ & 0.034 & $\backslash$ \\
& OURS & $\backslash$ & $\mathbf{0.033}$ & $\backslash$ \\
\hline 
\multirow{3}{*}{ degenerate\_01 } 
& LIO-SAM & $\backslash$ & 9.39 & $\backslash$ \\
& R3LIVE & $\backslash$ & $\mathbf{0.096}$ & $\backslash$ \\
& OURS & $\backslash$ & 5.18 & $\backslash$ \\
\hline 
\Xhline{2pt}
\end{tabular}
\end{table}
In this section, we compare the RMSE of the error $d_{p_{i}}$ of the point cloud registration before and after the degeneracy compensation, where $d_{p_{i}}$ can be obtained from (\ref{con:eqZ}). Furthermore, we conduct a comparison with three SLAM systems on the eight sequences.{The comparative methods are R3LIVE \cite{r14}, LIO-SAM \cite{r27}, and OURS. Among them, R3LIVE stands out as a robust SLAM system that integrates various sensors such as cameras, LiDAR, and IMU, harnessing their combined strengths for accurate localization and mapping; LIO-SAM utilizes Zhang J's method \cite{r10}, which is a lightweight LiDAR odometry; OURS is an algorithm that builds upon the foundation of LIO-SAM, replacing the degeneracy perception and resistance methods with the new approach proposed in this paper.} {All methods were implemented in C++ on ROS and executed on a laptop with an AMD R7-6800 CPU.} By conducting a comparative analysis of the positioning and mapping performance of R3LIVE, LIO-SAM, and OURS on these three sequences. This evaluation will help elucidate the efficacy of our proposed approach in mitigating degeneracy and improving SLAM performance.

As can be seen in Fig.\ref{fig11}, when LiDAR degeneracy occurs, the RMSE value of the point cloud registration becomes larger, and as can be seen in Fig.\ref{fig11b} the larger the degeneracy, the larger this RMSE value becomes. After our proposed method detects the degeneracy at that moment and uses the IMU to compensate for the LiDAR, the RMSE of degeneracy is significantly reduced after the compensation, except for the second degeneracy location in xhl, which was mentioned earlier as having experienced less degeneracy, demonstrates inferior results when IMU is hastily employed for LiDAR compensation. This demonstrates the effectiveness of our proposed degeneracy-aware and degeneracy-resistant algorithms in correctly recognizing degeneracy and effectively resisting LiDAR degeneracy, reducing the point cloud registration error at the moment of degeneracy.

{We compare OURS with R3LIVE and LIO-SAM, where all methods are versions without loop-closure. The comparison results are listed in Table \ref{tb2}. The sequences xhl, degenerate\_00, and degenerate\_01 have no ground truth, so we evaluate the algorithm using end-to-end error. OURS outperforms LIO-SAM for all eight sequences.  OURS and R3LIVE perform close in the campus, uzh, basement, and degenerate\_00 sequence. In xhl, quad, and construction OURS can perform overall better than R3LIVE and LIO-SAM. The camera introduced in R3LIVE degenerates in textureless environments, resulting in poorer performance for R3LIVE. LIO-SAM performs poorly in all eight sequences. The evaluation results highlight the limitations of LIO-SAM, which experiences severe errors and drift. The overall results under these degeneracy sequences show that our proposed degeneracy sensing and resistance method is effective and that OURS has higher localization accuracy and robustness.}

{Fig.\ref{fig12} shows the trajectories of the different methods on the eight degenerate sequences. The results show that the trajectory estimated by OURS and R3LIVE on the vast majority of sequences is satisfactory, while the results of LIO-SAM on all sequences are unsatisfactory. LIO-SAM undergoes severe degeneracy, and the trajectories all show significant errors, In particular, in the degenerate\_01 and construction sequences, where the actual shape of the trajectory is no longer visible. 
} 

\begin{figure*}[htbp]
\centering
\subfloat[]{\includegraphics[page=2, width=1.8in,height=3.6in]{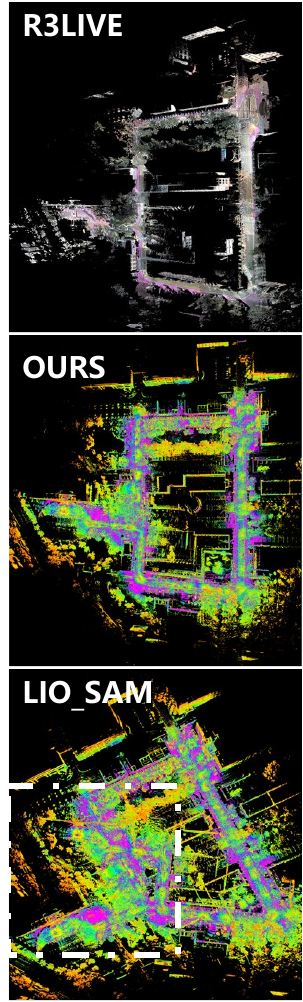}%
\label{fig13a}}
\subfloat[]{\includegraphics[page=3,width=1.8in,height=3.6in]{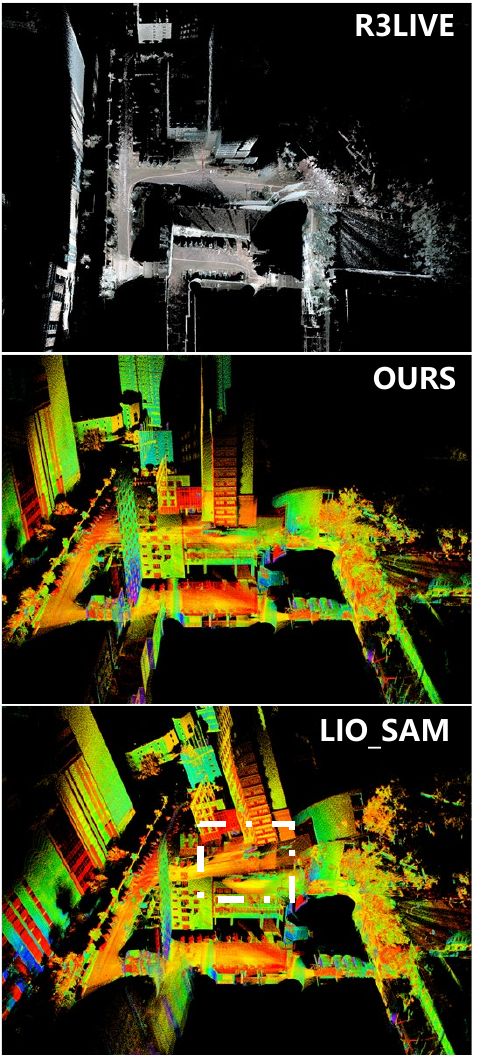}%
\label{fig13b}}
\subfloat[]{\includegraphics[page=6,width=1.8in,height=3.6in]{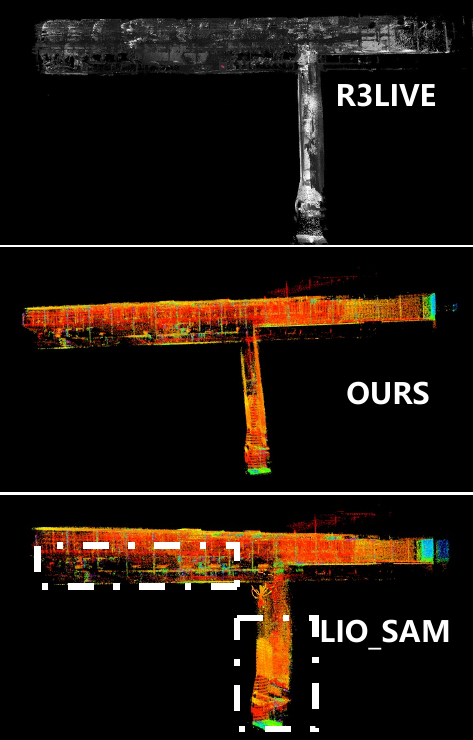}%
\label{fig13c}} 
\subfloat[]{\includegraphics[page=7, width=1.8in,height=3.6in]{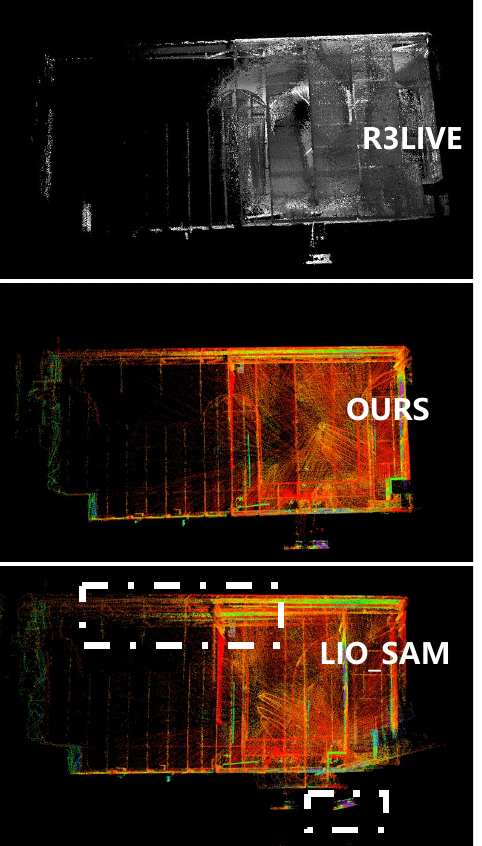}%
\label{fig13d}} \\
\subfloat[]{\includegraphics[page=8,width=1.8in,height=3.6in]{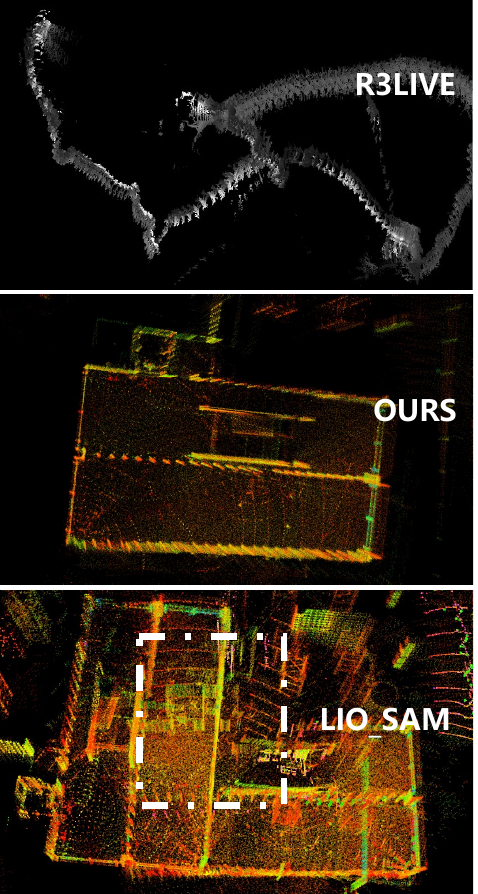}%
\label{fig13e}}
\subfloat[]{\includegraphics[page=9,width=1.8in,height=3.6in]{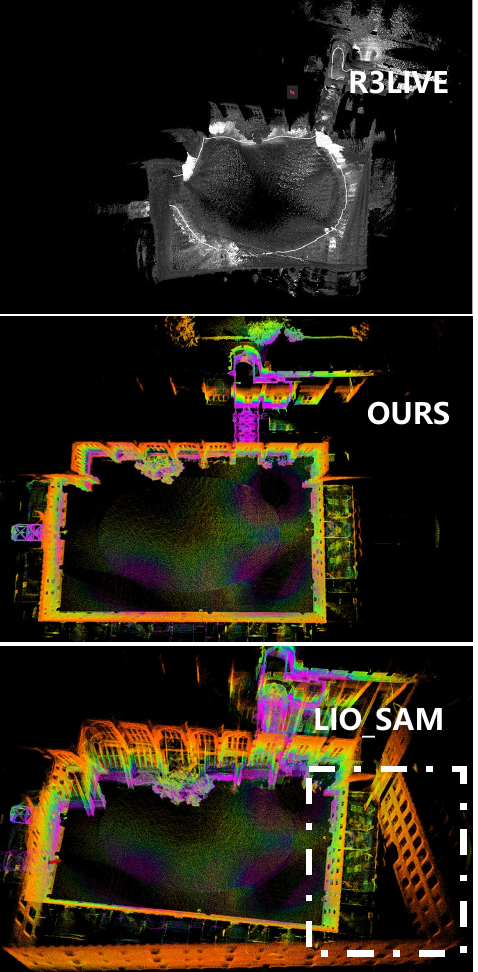}%
\label{fig13f}}
\subfloat[]{\includegraphics[page=4,width=1.8in,height=3.6in]{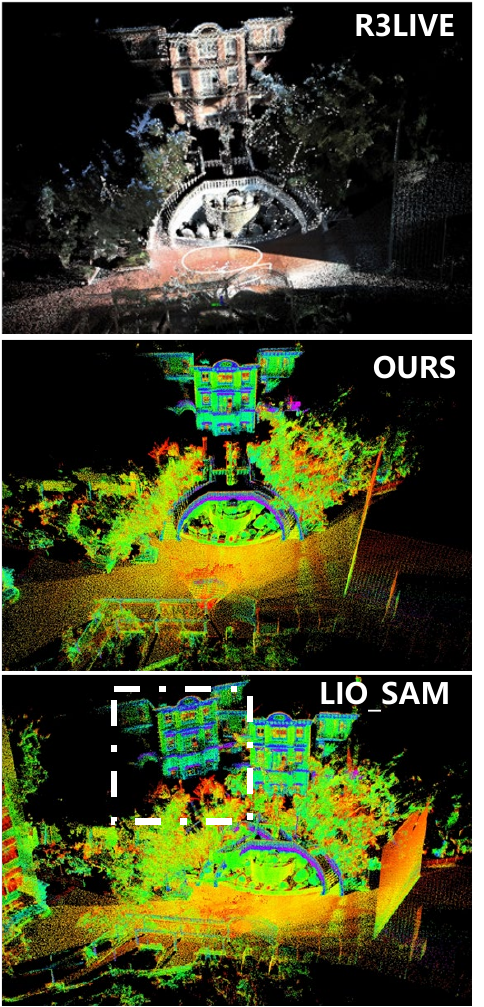}%
\label{fig13g}}
\subfloat[]{\includegraphics[page=5,width=1.8in,height=3.6in]{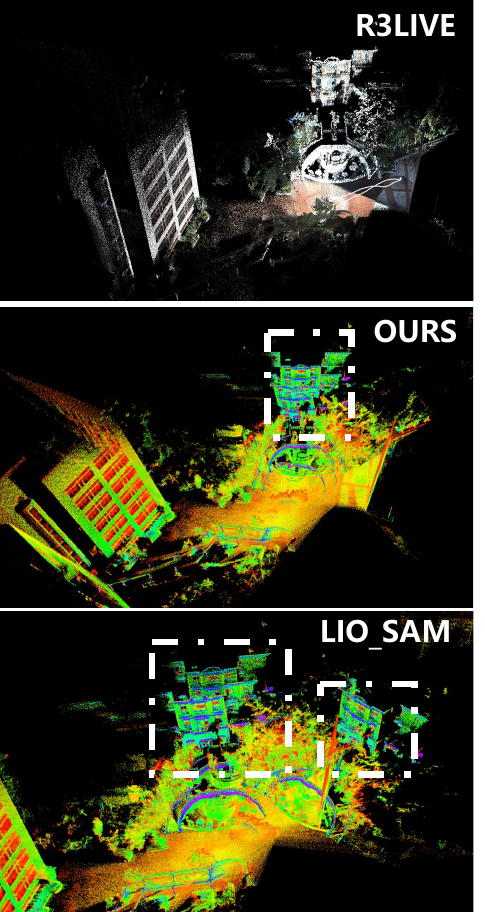}%
\label{fig13h}}
\caption{{The mapping results. (a)-(h) denote, respectively, campus, xhl, basement, uzh, quad, construction, and degenerate\_00, degenerate\_01. The white dashed box indicates a visible ghosting of the map.}} 
\label{fig13}
\end{figure*}

{Fig.\ref{fig13} illustrates the mapping results of R3LIVE, LIO-SAM, and OURS in degenerate sequences. In all point cloud maps generated by LIO-SAM, a clear ghosting effect is observable. By introducing a method for degeneracy perception and resistance, OURS has successfully addressed the ghosting issue of LIO-SAM, achieving consistent mapping results. Satisfactory mapping outcomes were obtained in all sequences except for the degenerate\_01 sequence.}

\section{Conclusion}
This paper proposes a method for real-time sensing and resistance of LiDAR degeneracy, improving the robustness of LiDAR positioning and mapping by dealing with the degeneracy of the environment. The degeneracy factor, defined and derived using the condition number, has a clear and distinct meaning. Moreover, the larger the value of this degeneracy factor, the more severe the degeneration of the LiDAR system becomes. At the same time, this paper adopts the DBSCAN clustering method to sense the degeneracy of LiDAR, abandons the traditional method of setting the threshold of degeneracy detection, and is not affected by the external environment and the type of LiDAR. When degeneracy occurs, the motion states of LiDAR and IMU are weighted and fused to reduce the impact of environmental degeneracy on LiDAR state estimation. We have used different types of LiDAR in multiple challenging scenarios for experimental verification. The experimental results show that the system can accurately perceive environmental degeneracy and measure the degree of LiDAR degeneracy. It significantly improves the state estimation of LiDAR in degenerate scenarios and obtains accurate positioning and mapping effects.

{However, our method still has some limitations. It is currently only suitable for sudden degeneracy, but our method is temporarily not applicable when LiDAR degenerates for a long time.} Because it is not an outlier when degeneracy occurs at this time but a new cluster, this requires us to divide the classification of DBSCAN in more detail, rather than simply dividing it into outliers and non-outliers, so that we can find the moment of LiDAR degeneracy more smoothly. This is also what we plan to do next.

\section*{Supplementary Material}
A video showing the result of positioning and mapping on the campus dataset is at \href{https://youtu.be/Q-I42J-lr-U}{https://youtu.be/Q-I42J-lr-U}.


\begin{IEEEbiography}[{\includegraphics[page=1, width=1in]{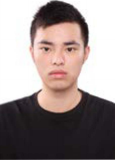}}]{Zongbo Liao} received the B.S. degree in 2022 from
the Electronic Information  School, Wuhan University. He is pursuing a Master's degree with the State Key Laboratory of Information Engineering in Surveying, Mapping, and Remote Sensing at Wuhan University, Wuhan, China. The main research direction is multi-sensor integrated navigation and digital signal processing.
\end{IEEEbiography}
\begin{IEEEbiography}[{\includegraphics[page=2, width=1in]{peo.pdf}}]{Xuanxuan Zhang}
obtained the B.S degree at the China University of Geosciences (Beijing) and the Master’s degree at the Chinese Academy of Surveying and Mapping. He is currently working toward a doctoral degree with the State Key Laboratory of Information Engineering in Surveying, Mapping, and Remote Sensing at Wuhan University, Wuhan, China. The main research direction is multi-sensor integrated navigation and GNSS high-precision data processing.
\end{IEEEbiography}
\begin{IEEEbiography}[{\includegraphics[page=3, width=1in]{peo.pdf}}]{Tianxiang Zhang}
 is pursuing a Master's degree at the State Key Laboratory of Surveying and Mapping Remote Sensing Information Engineering at Wuhan University. He was awarded the National First Prize of Intel Cup Undergraduate Electronic Design Contest - Embedded System Design Invitational Contest and designated as Outstanding Winner and AMS Award in the Interdisciplinary Contest in Modeling in 2022. His main research direction is Multi-sensor Simultaneous Localization and Mapping.
\end{IEEEbiography}
\begin{IEEEbiography}[{\includegraphics[page=4, width=1in]{peo.pdf}}]{Zhi Li}
is an engineer at the Beijing Automation Control Equipment Research Institute, affiliated with the CASIC, Beijing, North China. The main research directions are Visual navigation and multi-sensor integrated navigation.
\end{IEEEbiography}
\begin{IEEEbiography}[{\includegraphics[page=5, width=1in]{peo.pdf}}]{Zhenqi Zheng}
is pursuing a Ph.D. with the State Key Laboratory of Information Engineering in Surveying, Mapping, and Remote Sensing at Wuhan University, Wuhan, China. His main research interests include indoor positioning and navigation.
\end{IEEEbiography}
\begin{IEEEbiography}[{\includegraphics[page=6, width=1in]{peo.pdf}}]{Zhichao Wen}
is pursuing a Master's degree with the State Key Laboratory of Information Engineering in Surveying, Mapping, and Remote Sensing at Wuhan University, Wuhan, China. The main research direction is multi-sensor integrated navigation.
\end{IEEEbiography}
\begin{IEEEbiography}[{\includegraphics[page=7, width=1in]{peo.pdf}}]{You Li}
(Senior Member, IEEE) is a Professor at the State Key Laboratory of Information Engineering in Surveying, Mapping and Remote Sensing (LIESMARS), Wuhan University, China. He is also an Adjunct Professor at the Hubei Luojia Laboratory. He received Ph.D. degrees from Wuhan University and the University of Calgary in 2015 and 2016, respectively. His research focuses on positioning and motion-tracking techniques and applications. He has co-published over 100 academic papers and has over 20 patents. He serves as an Associate Editor for the IEEE SENSJ, a committee member at the IAG unmanned navigation system session, China Society of Surveying and Mapping location services session, and the secretary at the ISPRS mobile mapping session, and co-chairs of multiple international conferences, such as IROS and IPIN.
\end{IEEEbiography}

\vfill


\begin{thebibliography}{33}
\bibliographystyle{IEEEtran}





\bibitem{chiang2023performance}
K.~Chiang, Y.~Chiu, S.~Srinara, and M.~Tsai, ``Performance of lidar-slam-based pnt with initial poses based on ndt scan matching algorithm,'' \emph{Satellite Navigation}, vol.~4, no.~1, p.~3, 2023.

\bibitem{lou2022review}
Y.~Lou, X.~Dai, X.~Gong, C.~Li, Y.~Qing, Y.~Liu, Y.~Peng, and S.~Gu, ``A review of real-time multi-gnss precise orbit determination based on the filter method. satell navig 3 (1): 1--15,'' 2022.

\bibitem{r5}
S.~Thrun, M.~Montemerlo, H.~Dahlkamp, D.~Stavens, A.~Aron, J.~Diebel, P.~Fong, J.~Gale, M.~Halpenny, G.~Hoffmann \emph{et~al.}, ``Stanley: The robot that won the darpa grand challenge,'' \emph{Journal of field Robotics}, vol.~23, no.~9, pp. 661--692, 2006.

\bibitem{r6}
C.~Urmson, J.~Anhalt, D.~Bagnell, C.~Baker, R.~Bittner, M.~Clark, J.~Dolan, D.~Duggins, T.~Galatali, C.~Geyer \emph{et~al.}, ``Autonomous driving in urban environments: Boss and the urban challenge,'' \emph{Journal of field Robotics}, vol.~25, no.~8, pp. 425--466, 2008.

\bibitem{r7}
J.~Behley and C.~Stachniss, ``Efficient surfel-based slam using 3d laser range data in urban environments.'' in \emph{Robotics: Science and Systems}, vol. 2018, 2018, p.~59.

\bibitem{r8}
W.~Zhen and S.~Scherer, ``Estimating the localizability in tunnel-like environments using lidar and uwb,'' in \emph{2019 International Conference on Robotics and Automation (ICRA)}.\hskip 1em plus 0.5em minus 0.4em\relax IEEE, 2019, pp. 4903--4908.

\bibitem{r9}
C.~Rizzo, D.~Tardioli, D.~Sicignano, L.~Riazuelo, J.~L. Villarroel, and L.~Montano, ``Signal-based deployment planning for robot teams in tunnel-like fading environments,'' \emph{The International Journal of Robotics Research}, vol.~32, no.~12, pp. 1381--1397, 2013.

\bibitem{r10}
J.~Zhang, M.~Kaess, and S.~Singh, ``On degeneracy of optimization-based state estimation problems,'' in \emph{2016 IEEE International Conference on Robotics and Automation (ICRA)}.\hskip 1em plus 0.5em minus 0.4em\relax IEEE, 2016, pp. 809--816.

\bibitem{r11}
J.~Lin, C.~Zheng, W.~Xu, and F.~Zhang, ``R2live: A robust, real-time, lidar-inertial-visual tightly-coupled state estimator and mapping. arxiv 2021,'' \emph{arXiv preprint arXiv:2102.12400}.

\bibitem{r12}
S.~Khattak, H.~Nguyen, F.~Mascarich, T.~Dang, and K.~Alexis, ``Complementary multi--modal sensor fusion for resilient robot pose estimation in subterranean environments,'' in \emph{2020 International Conference on Unmanned Aircraft Systems (ICUAS)}.\hskip 1em plus 0.5em minus 0.4em\relax IEEE, 2020, pp. 1024--1029.

\bibitem{r13}
T.~Shan, B.~Englot, C.~Ratti, and D.~Rus, ``Lvi-sam: Tightly-coupled lidar-visual-inertial odometry via smoothing and mapping,'' in \emph{2021 IEEE international conference on robotics and automation (ICRA)}.\hskip 1em plus 0.5em minus 0.4em\relax IEEE, 2021, pp. 5692--5698.

\bibitem{r14}
J.~Lin and F.~Zhang, ``R 3 live: A robust, real-time, rgb-colored, lidar-inertial-visual tightly-coupled state estimation and mapping package,'' in \emph{2022 International Conference on Robotics and Automation (ICRA)}.\hskip 1em plus 0.5em minus 0.4em\relax IEEE, 2022, pp. 10\,672--10\,678.

\bibitem{r15}
M.~Palieri, B.~Morrell, A.~Thakur, K.~Ebadi, J.~Nash, A.~Chatterjee, C.~Kanellakis, L.~Carlone, C.~Guaragnella, and A.-a. Agha-Mohammadi, ``Locus: A multi-sensor lidar-centric solution for high-precision odometry and 3d mapping in real-time,'' \emph{IEEE Robotics and Automation Letters}, vol.~6, no.~2, pp. 421--428, 2020.

\bibitem{r16}
W.~Zhen, S.~Zeng, and S.~Soberer, ``Robust localization and localizability estimation with a rotating laser scanner,'' in \emph{2017 IEEE international conference on robotics and automation (ICRA)}.\hskip 1em plus 0.5em minus 0.4em\relax IEEE, 2017, pp. 6240--6245.

\bibitem{koizumi2017avoidance}
M.~Koizumi, K.~Nonaka, and K.~Sekiguchi, ``Avoidance of singular localization environment using model predictive control for mobile robots,'' in \emph{2017 11th Asian Control Conference (ASCC)}.\hskip 1em plus 0.5em minus 0.4em\relax IEEE, 2017, pp. 2866--2871.

\bibitem{r21}
A.~Hinduja, B.-J. Ho, and M.~Kaess, ``Degeneracy-aware factors with applications to underwater slam,'' in \emph{2019 IEEE/RSJ International Conference on Intelligent Robots and Systems (IROS)}.\hskip 1em plus 0.5em minus 0.4em\relax IEEE, 2019, pp. 1293--1299.

\bibitem{rong2016detection}
Z.~Rong and N.~Michael, ``Detection and prediction of near-term state estimation degradation via online nonlinear observability analysis,'' in \emph{2016 IEEE International Symposium on Safety, Security, and Rescue Robotics (SSRR)}.\hskip 1em plus 0.5em minus 0.4em\relax IEEE, 2016, pp. 28--33.


\bibitem{r18}
H.~Cho, S.~Yeon, H.~Choi, and N.~Doh, ``Detection and compensation of degeneracy cases for imu-kinect integrated continuous slam with plane features,'' \emph{Sensors}, vol.~18, no.~4, p. 935, 2018.

\bibitem{ebadi2021dare}
K.~Ebadi, M.~Palieri, S.~Wood, C.~Padgett, and A.-a. Agha-mohammadi, ``Dare-slam: Degeneracy-aware and resilient loop closing in perceptually-degraded environments,'' \emph{Journal of Intelligent \& Robotic Systems}, vol. 102, pp. 1--25, 2021.









\bibitem{r24}
P.~D. Groves, ``Principles of gnss, inertial, and multisensor integrated navigation systems, [book review],'' \emph{IEEE Aerospace and Electronic Systems Magazine}, vol.~30, no.~2, pp. 26--27, 2015.

\bibitem{r25}
J.~Zhang and S.~Singh, ``Loam: Lidar odometry and mapping in real-time.'' in \emph{Robotics: Science and systems}, vol.~2, no.~9.\hskip 1em plus 0.5em minus 0.4em\relax Berkeley, CA, 2014, pp. 1--9.

\bibitem{r26}
H.~Ye, Y.~Chen, and M.~Liu, ``Tightly coupled 3d lidar inertial odometry and mapping,'' in \emph{2019 International Conference on Robotics and Automation (ICRA)}.\hskip 1em plus 0.5em minus 0.4em\relax IEEE, 2019, pp. 3144--3150.

\bibitem{r27}
T.~Shan, B.~Englot, D.~Meyers, W.~Wang, C.~Ratti, and D.~Rus, ``Lio-sam: Tightly-coupled lidar inertial odometry via smoothing and mapping,'' in \emph{2020 IEEE/RSJ international conference on intelligent robots and systems (IROS)}.\hskip 1em plus 0.5em minus 0.4em\relax IEEE, 2020, pp. 5135--5142.

\bibitem{r28}
C.~Qin, H.~Ye, C.~E. Pranata, J.~Han, S.~Zhang, and M.~Liu, ``Lins: A lidar-inertial state estimator for robust and efficient navigation,'' in \emph{2020 IEEE international conference on robotics and automation (ICRA)}.\hskip 1em plus 0.5em minus 0.4em\relax IEEE, 2020, pp. 8899--8906.

\bibitem{r29}
W.~Xu and F.~Zhang, ``Fast-lio: A fast, robust lidar-inertial odometry package by tightly-coupled iterated kalman filter,'' \emph{IEEE Robotics and Automation Letters}, vol.~6, no.~2, pp. 3317--3324, 2021.

\bibitem{r30}
C.~H. Tong, P.~Furgale, and T.~D. Barfoot, ``Gaussian process gauss--newton for non-parametric simultaneous localization and mapping,'' \emph{The International Journal of Robotics Research}, vol.~32, no.~5, pp. 507--525, 2013.

\bibitem{r22}
D.~A. Belsley, E.~Kuh, and R.~E. Welsch, \emph{Regression diagnostics: Identifying influential data and sources of collinearity}.\hskip 1em plus 0.5em minus 0.4em\relax John Wiley \& Sons, 2005.


\bibitem{r32}
M.~Ester, H.-P. Kriegel, J.~Sander, X.~Xu \emph{et~al.}, ``A density-based algorithm for discovering clusters in large spatial databases with noise,'' in \emph{kdd}, vol.~96, no.~34, 1996, pp. 226--231.

\bibitem{r33}
Helmberger M, Morin K, Berner B, \emph{et~al.}, ``The hilti slam challenge dataset,``. \emph{IEEE Robotics and Automation Letters}, vol.~7, no.~3, pp. 7518--7525, 2022.

\end{thebibliography}
\end{document}